\RequirePackage{fix-cm}

\documentclass[smallextended]{svjour3}

\smartqed  

\usepackage{tabularx}

\usepackage{graphicx}

\usepackage[authoryear]{natbib}  

\usepackage[resetlabels]{multibib}
\newcites{inappendix}{Supplementary References}

\usepackage{hyperref}

\usepackage{subcaption}

\usepackage{hhline}

\usepackage{amsmath}

\usepackage{rotating}

\usepackage{bibspacing}
\ifdefined\mydraft
\mydraft
\fi

\graphicspath{{img/}}

\title{Matchmaker, Matchmaker, Make Me a Match: Geometric, Variational, and Evolutionary Implications of Criteria for Tag Affinity}
\titlerunning{Matchmaker, Matchmaker, Make Me a Match}
\author{
    Matthew Andres Moreno
	  \and Alexander Lalejini
    \and Charles Ofria \\
}

\authorrunning{Moreno et al.} 

\institute{M. A. Moreno \at
  BEACON Center for the Study of Evolution in Action \\
  Department of Computer Science and Engineering \\
  Program in Ecology, Evolutionary Biology, and Behavior \\
  Michigan State University, East Lansing, MI \\
  \email{mmore500@msu.edu}
\and
  A. Lalejini \at
  University of Michigan, Ann Arbor, MI
\and
  C. Ofria \at
  BEACON Center for the Study of Evolution in Action \\
  Department of Computer Science and Engineering \\
  Program in Ecology, Evolutionary Biology, and Behavior \\
  Michigan State University, East Lansing, MI \\
}

\date{Received: date / Accepted: date}

\begin{document}
\maketitle

\begin{abstract}

Genetic programming and artificial life systems commonly employ tag-matching schemes to determine interactions between model components.
However, the implications of criteria used to determine affinity between tags with respect to constraints on emergent connectivity, canalization of changes to connectivity under mutation, and evolutionary dynamics have not been considered.
We highlight differences between tag-matching criteria with respect to geometric constraint and variation generated under mutation. 
We find that tag-matching criteria can influence the rate of adaptive evolution and the quality of evolved solutions.
Better understanding of the geometric, variational, and evolutionary properties of tag-matching criteria will facilitate more effective incorporation of tag matching into genetic programming and artificial life systems.
By showing that tag-matching criteria influence connectivity patterns and evolutionary dynamics, our findings also raise fundamental questions about the properties of tag-matching systems in nature.


\end{abstract}
 
\section{Introduction}

Computer operations must specify the identities of registers and memory addresses they read from and act on.
Computer programs, composed of these individual operations, apply identity-specifying information in individual operations to name higher-level abstractions such as data structures and functions.
As instances of computer programs by their very nature, genetic programs and digital artificial life systems must also specify computational elements on which to act.
These computational elements range from
\begin{itemize}
  \item program modules in genetic programs \citep{spector2011tag}
  \item virtual hardware analogs like registers, memory addresses, stacks, or jump addresses in genetic programs \citep{lalejini_tag-accessed_2019,ray1991approach,ofria2004avida},
  \item molecules in artificial chemistries \citep{bagley1990spontaneous},
  \item genes in artificial gene regulatory networks \citep{banzhaf2003artificial},
  \item individual neurons or neural modules in neuroevolution \citep{reisinger2007acquiring}, or 
  \item agents in agent-based models of complex systems \citep{riolo2001evolution}.
\end{itemize}

Many artificial life systems are conceived to study open-ended dynamics.
As such, it is often essential to allow for the computational elements designated for particular operations to shift over time, to allow for the introduction and incorporation of novel computational objects at run time, and to allow for the removal of existing computational objects at run time.
Likewise, dynamic reorganization of code modules can facilitate hierarchical problem-solving in genetic programming \citep{Kinnear:Koza:1994:adf}.

Tag-based referencing, sometimes also termed pattern matching or inexact referencing, provides a practical and commonplace solution for specifying computational operands.
This approach encodes a tag for each computational operand that may be selected and a tag for each querying operation.
Operands are then selected for each query through a tag-matching process.
A querying operation's tag is compared to available operand tags.
Then, typically, either:
\begin{itemize}
  \item the best-matching operand is selected (e.g., \citep{spector2012tag}),
  \item all operands with match quality exceeding a threshold are selected (e.g., \citep{riolo2001evolution}),
  \item operands are activated to continuously-varying degrees based on match qualities (e.g., \citep{banzhaf2003artificial}), or
  \item operands are selected probabilistically based on match quality (e.g., \citep{seiden1992simulation}).
\end{itemize}

Inexact referencing facilitates orderly growth, shrinkage, and reconfiguration of a system's set of operands and operations.
If an operand is deleted, it does not invalidate any existing operations, as other well-matching operands will fill its place.
Likewise, new operations can be created or existing operations can be altered freely because arbitrary query tags may select operands. 

Indeed, inexact referencing techniques find common use in agent-based modeling \citep{riolo2001evolution}, neuroevolution \citep{reisinger2007acquiring}, artificial gene regulatory networks \citep{banzhaf2003artificial}, genetic programming \citep{spector2011tag, lalejini2018evolving}, artificial chemistry \citep{dittrich2001artificial}, and artificial immunology \citep{timmis2008theoretical}.
These systems typically either use tagging schemes based on
\begin{itemize}
    \item on hamming distance between bitstrings (\textit{e.g.}, \cite{lalejini2018evolving} or \cite{banzhaf2003artificial}) or 
    \item differences between real-valued scalars (\textit{e.g.}, \cite{riolo2001evolution} or \cite{spector2011tag}).
\end{itemize}    
However, other nonlinear tag-matching systems, such as Downing's streak metric \citep{downing2015intelligence} and de Boer's adjacency match metric \citep{DEBOER1991381}, have been proposed.

Although some efforts have been made to distinguish certain tag-matching criteria in terms of narrative explanations of their evolutionary properties and appeals to biological analogy \citep{downing2015intelligence,scherer2004activation}, no work has yet provided systematic, quantitative, or empirical insight into the ramifications of tag-matching criteria.
We hypothesize that properties of tag-matching criteria could potentially affect evolvability through mechanisms including
\begin{itemize}
  \item bias of certain queries or operands against tight-affinity matches (i.e., tunable specificity)
  \item bias to the stability of certain connections under mutation (i.e., tunable robustness),
  \item bias to the likelihood of connections arising between subsets of queries and operands (i.e., modularity), and
  \item mitigation of disruption under duplication of queries and operands (i.e., gene duplication \citep{ohno2013evolution, lewis1978gene}).
\end{itemize}

In this work, we show how tag-matching systems vary with respect to
\begin{enumerate}
  \item geometric structure that biases or limits the patterns of connectivity that form among queries and operands (Section \ref{sec:geometric}),
  \item variational properties that influence changes to connectivity observed under mutation (Section \ref{sec:variational}), and
  \item evolutionary outcomes such as the rate of adaptive evolution and the quality of evolved solutions (Section \ref{sec:evolutionary}.
\end{enumerate}
Understanding these differences will help researchers more effectively employ tag-matching schemes in artificial life systems, but also provide context for inquiry into the properties and mechanisms of tag-matching systems in nature.

\section{Tags and Tag-Matching Metrics}

\begin{figure}
\begin{center}

\includegraphics[width=\columnwidth]{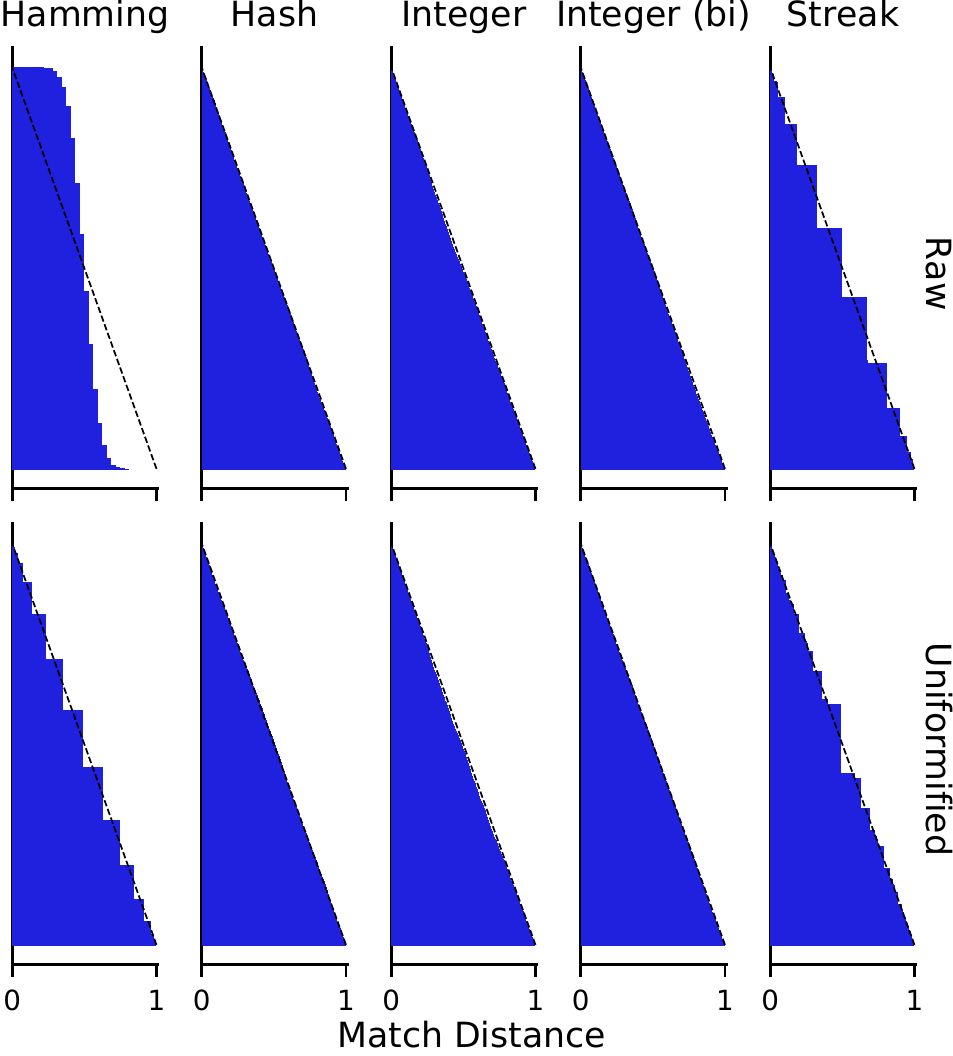}
\caption{
Distance distributions of metrics before and after uniformification.
Dashed line indicates an ideal uniform distribution.
}
\label{fig:uniformification}

\end{center}
\end{figure}
 
In all experiments, we used 32-bit bitstrings as tags.
Formally, we define a tag $t$ as a fixed-length binary vector,
\begin{align*}
t = \langle t_0, t_1, t_2, \dots, t_{n-2}, t_{n-1} \rangle
\end{align*}
where
\begin{align*}
\forall i, t_i \in \{0, 1\} \text{ and } n=32.
\end{align*}

In experiments where mutations were applied to tags, individual bits were toggled stochastically at a uniform per-bit rate.

We call an algorithm used to calculate the match quality between two tags a tag-matching metric.
A tag-matching metric takes two tags as operands and calculates a match distance between them.
Low match distance indicates a ``good'' or ``strong'' match.
High match distance indicates a ``poor'' or ``weak'' match.

We compared five tag-matching metrics: hamming, hash, integer, bidirectional integer, and streak.
The hamming and bidirectional integer metrics are included because of their ubiquity in artificial life systems.
The integer metric is included due to its use in seminal work exploring tag-matching in genetic programs \citep{spector2011tag, spector2011s,spector2012tag}.
The streak metric was proposed to model large-effect mutations in biology but, to our knowledge, has not yet been formally studied in an evolving system.
The hash metric is introduced in this work in order to investigate the implications of a completely geometrically-unstructured tag-matching scheme.
Table \ref{tab:metrics} compares summary descriptions for each metric.

\begin{table}[]
\begin{tabularx}{\textwidth}{l|X}
\textbf{Metric}       & \textbf{Description}                                                                                                                                        \\ \hline
Hamming               & fraction of positions within \texttt{tag\_0} and \texttt{tag\_1} with mismatching bits                                                                         \\ \hline
Hash                  & SHA1 cryptographic hash of  concatenation of \texttt{tag\_0} and \texttt{tag\_1} \citep{eastlake2001us}                         \\ \hline
Integer               & value added to the unsigned integer representation of \texttt{tag\_0} to reach representation of \texttt{tag\_1}, wrapping around if necessary \\ \hline
Bidirectional Integer & lesser of integer metric distances \texttt{d(tag\_0, tag\_1)} and \texttt{d(tag\_1, tag\_0)}                                                                \\ \hline
Streak                & ratio of lengths of contiguously matching and mismatching substrings \\ \hline
\end{tabularx}

\begin{tabularx}{\textwidth}{l|X|X}
\hline
\textbf{Metric}       & \textbf{Commutative?} & \textbf{Multidimensional?} \\ \hline
Hamming               & yes                   & yes                        \\ \hline
Hash                  & no                    & yes                        \\ \hline
Integer               & no                    & no                         \\ \hline
Bidirectional Integer & yes                   & no                         \\ \hline
Streak                & yes                   & yes                       
\end{tabularx}

\caption{
Surveyed tag-matching metrics. A matching metric is commutative if \texttt{d(tag\_0, tag\_1) = d(tag\_1, tag\_0)} for all tags.
A matching metric is considered multidimensional if position within matching space is not represented by a scalar value. 
}
\label{tab:metrics}

\end{table}

Sections \ref{sec:hamming}, \ref{sec:hash}, \ref{sec:integer}, \ref{sec:bidirectionalinteger}, and \ref{sec:streak} provide formal definitions for each metric.

\subsection{Hamming Metric} \label{sec:hamming}

The hamming metric computes match distance as the fraction of positions between tags $t$ and $u$ with mismatching bits.
Formally, for $n$-bit bitstring tags,
\begin{align*}
d(t, u)
= \frac{
  \#\{ i : t_i \neq u_i, i=0, \dots ,n-1\}
}{
  n
}.
\end{align*}

This metric is based on \cite{lalejini2019else}, originally after \cite{hamming1950error}.

\subsection{Hash Metric} \label{sec:hash}

The hash metric calculates match distance via a cryptographic hash of tags $t$ and $u$.
First, we \texttt{memcpy} $t$ and $u$ into a double-width bitstring $v$ such that
\begin{align*}
v = \langle t_0, t_1, t_2, \dots, t_{n-2}, t_{n-1}, u_0, u_1, u_2, \dots, u_{n-2}, u_{n-1} \rangle
\end{align*}

Then, we use the OpenSSL library to generate a \texttt{std::string} digest of $v$.
We then apply \texttt{std::hash} to map this digest to a \texttt{std::size\_t}, $v'$.
Finally, we compute the matching distance as
\begin{align*}
d(t, u) = \frac{v'}{\texttt{static\_cast<double>(std::numeric\_limits<std::size\_t>::max())}}.
\end{align*}

To our knowledge, the hash metric is original to this work.
The metric produces an arbitrary, but deterministic, match distance between any two tags.
In other words, the tag matching space is completely unstructured.
We include it primarily to serve as a control.

\subsection{Integer Metric} \label{sec:integer}

The integer metric computes match distance between tags $t$ and $u$ by counting upwards from $t$ until $u$ is reached.
If necessary, the counting process wraps around at $2^n$.

To accomplish this, the integer metric must interpret bitstring tags $t$ and $u$ as unsigned integers.
We use a standard representation,
\begin{align*}
f(t)
= \sum_{i=0}^{n-1} t_i \times 2^i.
\end{align*}

Formally, the integer metric computes distance between $n$-bit bitstring tags as,
\begin{align*}
d(t, u) =
\begin{cases}
  \frac{2^n - f(t) + f(u)}{2^n}, & \text{if } f(t) > f(u), \\
  \frac{f(t) - f(u)}{2^n},         & \text{otherwise}.
\end{cases}
\end{align*}

Inclusion of this metric is motivated by \cite{spector2011tag}, who used positive integers between 0 and 100 to name referents.
Queries matched to the referent that had the next-larger value, wrapping around from 100 back to 0.

\section{Bidirectional Integer Metric} \label{sec:bidirectionalinteger}

The bidirectional integer metric computes match distance between tags $t$ and $u$ by counting from $t$ to $u$.
The count from $t$ to $u$ ascends or descends, whichever is shorter.
If necessary, the count wraps around at $0$ and $2^n$.

The binteger metric interprets bitstring tags $t$ and $u$ as unsigned integers using the same mapping, $f$, as the integer metric.

Formally, the bidirectional integer metric computes distance between $n$-bit bitstring tags as,
\begin{align*}
d(t, u) =
\min
\begin{cases}
  \frac{2^n - \max\big(f(t), f(u)\big) + \min\big(f(t), f(u)\big)}{2^n}, \\
  \frac{\max\big(f(t), f(u)\big) - \min\big(f(t), f(u)\big)}{2^n}.
\end{cases}
\end{align*}

We included this metric to contrast with the integer metric. In particular, we wished to shed light on any consequences of its asymmetry and discontinuity.
In figure axes and legends with tight space constraints, we refer to this metric as ``Integer (bi)''.

\subsection{Streak Metric} \label{sec:streak}

The streak metric computes match distance between bitstring tags $t$ and $u$ as a ratio of lengths of contiguously matching and mismatching substrings within those tags.

Formally, we can compute the greatest contiguously-matching length of $n$-long bitstrings $t$ and $u$ as,
\begin{align*}
m(t, u) = \max\Big(\{i - j \forall i, j \in 0..n-1 \mid \forall q \in i..j, t_q = u_q \}\Big).
\end{align*}

Likewise, the greatest contiguously-mismatching length can be computed as,
\begin{align*}
n(t, u) = \max\Big(\{i - j \forall i, j \in 0..n-1 \mid \forall q \in i..j, t_q \neq u_q \}\Big).
\end{align*}

As proposed in \cite{downing2015intelligence}, the streak metric computes distance between $n$-bit bitstring tags $t$ and $u$ as,
\begin{align*}
d'(t, u)
= \frac{p'(n(t,u))}{p'(m(t,u)) + p'(n(t,u))}.
\end{align*}
where $p(k)$ approximates the probability of a contiguous $k$-bit match between two bitstrings.
\cite{downing2015intelligence} derives
\begin{align*}
p'(k)
= \frac{n - k + 1}{2^k}.
\end{align*}

However, this formula is subtly flawed.
For instance, the probability of a $0$-bit match according to this formula would be computed as $p'(0) = \frac{n - 0 + 1}{2^0} = n + 1$.
This is clearly impossible --- it would imply $p'(0) > 1 \forall n > 0$.

Although correct probabilities can be calculated via dynamic programming, $p'$ provides a useful approximation.
For computational efficiency and consistency with the existing literature, we use the math proposed in \citep{downing2015intelligence} but clamp edge cases between 0.0 and 1.0.
This yields the corrected streak metric $d$ used in this work,
\begin{align*}
d(t, u) = \max\Big( \min( d'(t, u), 1), 0 \Big).
\end{align*}

Downing's presentation of the streak metric motivates it by analogy to the biochemistry of enzyme biochemistry. 
In motivating the metric, Downing reports mutational walk experiments that show it to exhibit greater robustness compared to integer and hamming metrics.
However, it is not demonstrated in an evolving system.
To our knowledge, no further work on this metric has been published.
(Although, through personal communication, we learned of some unpublished work applying the metric in a neuroevolution system.)

\subsection{Match Distance Normalization}

For consistency of implementation and interpretation, all metrics' formulas return tag-matching distances between 0.0 (a ``perfect'' match) and 1.0 (a ``worst'' match).

However, the distribution of tag-match distances within this range may vary substantially between metrics.
For example, the probability of a match distance <1/32 is 1/32 under the hash metric but $1/2^32$ under the hamming metric.

In order to ensure an intuitive interpretation of match distances that was consistent across all tag-matching metrics, we normalized metrics' match distances so that the distances between pairs of randomly generated tags would follow a uniform distribution between 0.0 and 1.0.
In this discussion, we refer to match distance before normalization as ``raw.''

For example, two tags with a 0.01 match distance are better-matched than 99\% of randomly-generated tag pairs.
Additionally, in situations where raw match distance plays a mechanistic role (for example, probabilistic matching or threshold-based cutoffs), this transformation ensures consistency across metrics.

We performed this normalization independently for each tag-matching metric.
We the following Monte Carlo approximation method.
\begin{enumerate}
\item We sampled 10,000 pairs of randomly-generated tags.
\item We calculated raw match distance between each pair of generated tags using the chosen tag-matching metric.
\item We agglomerated these 10,000 sampled raw match distances into a list and sorted in ascending order.
\item To ensure coverage of the entire $[0.0,1.0]$ interval of valid tag match scores, we prepended the sorted list of raw match distances with 0.0 and 1.0.
\item We associated each list entry with its percentile ranking within the list.
\begin{enumerate}
    \item i.e., the best-matching 0.0 match distance was associated with the percentile ranking 0.0,
    \item the median match distance was associated with the percentile ranking 0.5, and
    \item the worst-matching 1.0 match distance was associated with the percentile ranking 1.0.
\end{enumerate}
\item For subsequent tag match distance calculations during the experiment, we performed a lookup on this list.
\begin{itemize}
    \item If a single exactly-identical raw match distance existed in the list, we returned its percentile ranking as the normalized match distance.
    \item If two or more exactly-identical raw match distances existed in the list, we returned the mean percentile ranking of these entries as the normalized match distance.
    \item If no exactly-identical raw match distance existed in the list, we linearly interpolated between the next-largest and next-smallest list entries' percentile rankings. 
\end{itemize}
\end{enumerate}
Figure \ref{fig:uniformification} compares the distribution of match distances between randomly-sampled tags before and after this normalization process across tag-matching metrics.

All work reported here employed match distance normalization. 
\section{Geometric Analyses} \label{sec:geometric}

In this section, we consider the geometry that tag matching metrics impose over bitstring tag space.
These geometries may affect the patterns of connectivity between tagged components that tend to (or even are possible to) arise.

As an example of potential geometric constraint, consider the bitstring tags $t = \langle 0, 0, ... 0 \rangle$ and $u = \langle 1, 1, ... 1 \rangle$ under the hamming metric.
No third tag $v$ could simultaneously exhibit a tag match distance < 0.5 to both tags.
However, under the hash metric no such pair of tags exists --- how well a third tag $v$ matches to $t$ and how well it matches to $u$ is always entirely independent.
Stated more generally, in metrics with strong geometric constraints, there may commonly exist pairs of tags such that no single third tag can simultaneously exhibit a close affinity to both.

As another example of potential geometric constraint, consider the bitstring tags $t = \langle 0, 0, ... 0 \rangle$ and $u = \langle 0, 0, ... 1 , 0 \rangle$ under the bidirectional integer metric.
(Here, the tag $t$ would correspond to the integer 0 and the tag $u$ would correspond to the integer 1.)
No third tag $v$ could simultaneously exhibit a match distance > 0.9 to $t$ and < 0.1 to $u$.
However, under the integer metric the $v = 1$ would satisfy these criteria.
Stated more generally, in metrics with strong geometric constraints, there may commonly exist pairs of tags such that any third tag must either match both closely or match neither closely.

Geometric constraint seems likely to profoundly influence evolution in tag-matching systems.
However, understanding how these implications ultimately play out is a difficult problem.
Geometric constraint might prove useful to facilitate modularity, where subsets of tag space tend to have associated functionality \citep{holland1990concerning}.
However, it may also restrict the generation of adaptive variation.

We begin by comparing distributions of two statistics measuring constraint across our five tag-matching metrics: similarity constraint and dissimilarity constraint.
\textit{Similarity constraint}, presented in Section \ref{sec:dissimilarityconstraint} quantifies the question, "If two tags both match closely to a third tag, will they necessarily match closely with each other?"
In contrast, \textit{dissimilarity constraint} quantifies the question, ``If a certain tag matches a second tag closely and a third tag poorly, will the second and third tag tend to match poorly?"

Finally, in Section

\subsection{Similarity Constraint} \label{sec:similarityconstriant}

\begin{figure*}
\begin{center}

\begin{minipage}{\linewidth}
\begin{subfigure}[b]{\linewidth}
\begin{minipage}{0.5\textwidth}
\begin{center}
\includegraphics[width=0.5\linewidth,trim=5cm 5cm 5cm 5cm, clip]{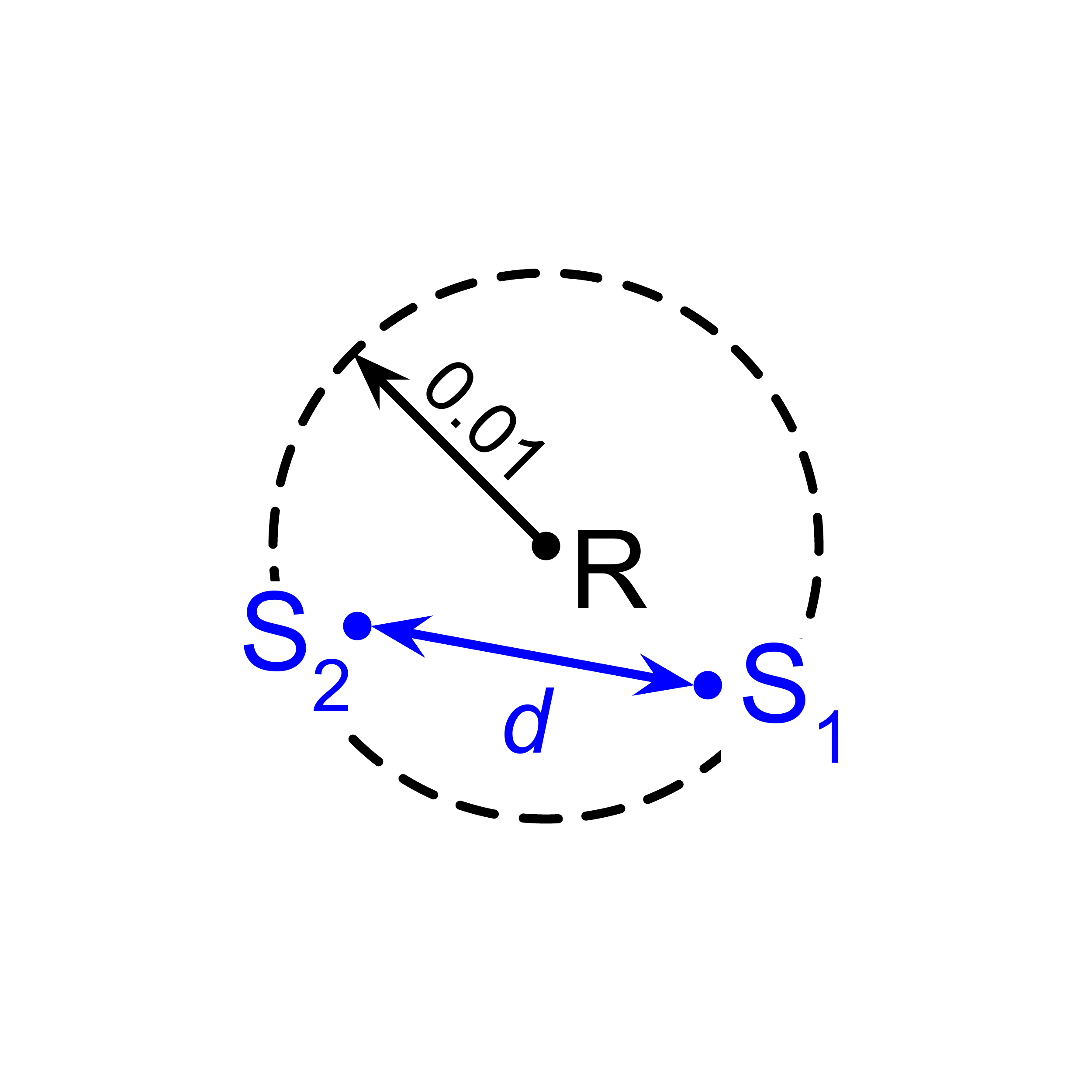}
\end{center}
\end{minipage}\begin{minipage}{0.5\textwidth}
\caption{
Sampling process used to evaluate similarity constraint.
}
\label{fig:dimensionality_measure}
\end{minipage}
\end{subfigure}
\end{minipage}
\begin{subfigure}[b]{\linewidth}
\begin{minipage}{0.6\linewidth}
\includegraphics[width=\linewidth]{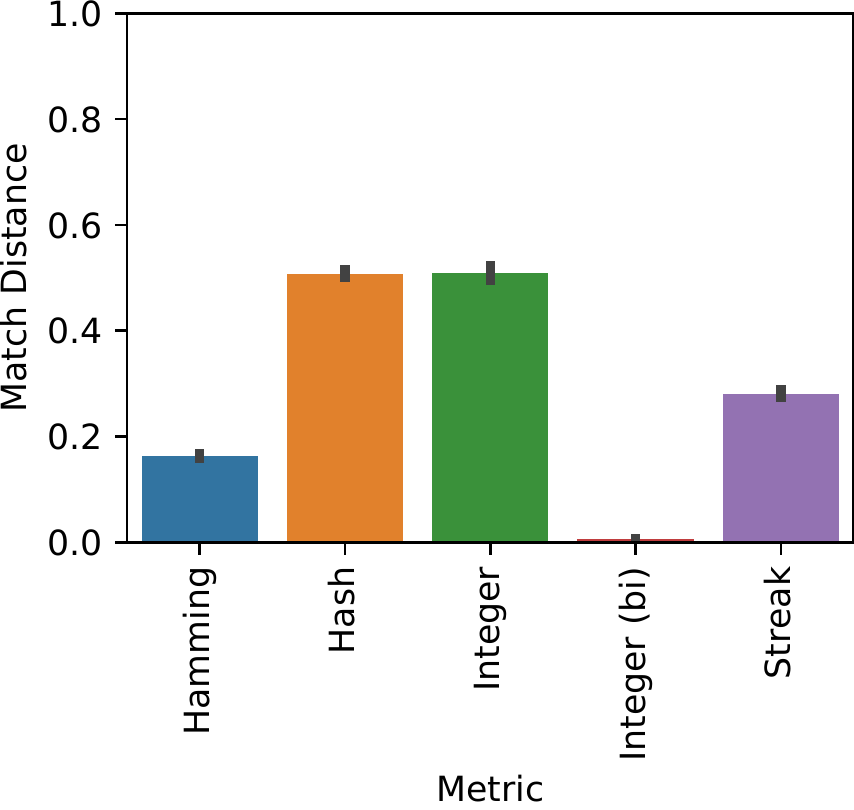}
\end{minipage}
\begin{minipage}{0.35\linewidth}
\caption{
Mean similarity constraint.
Error bars represent 95\% confidence intervals.
}
\label{fig:sphere_barplot}
\end{minipage}
\end{subfigure}
\begin{minipage}{\linewidth}
\begin{subfigure}[b]{\linewidth}
\centering
\includegraphics[width=\linewidth]{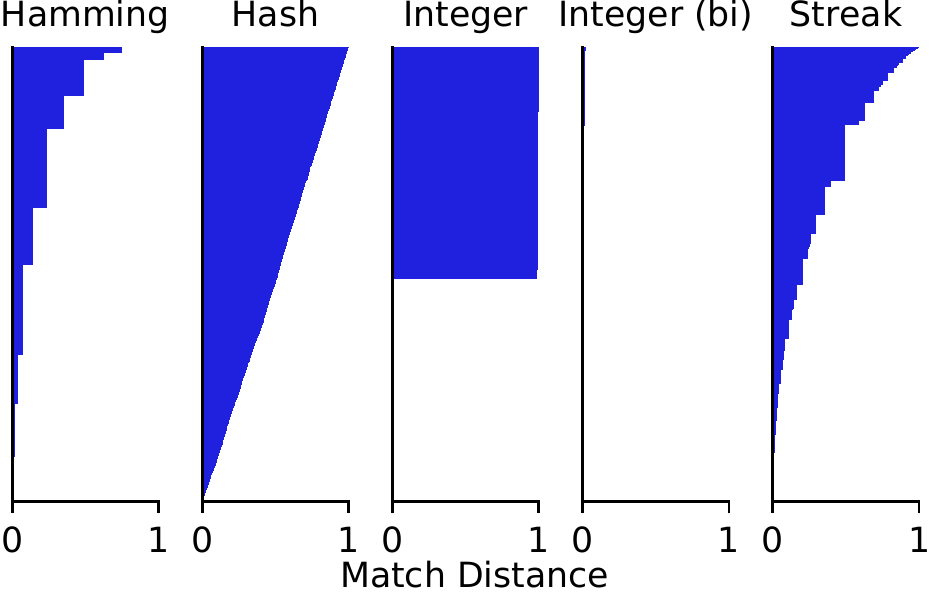}
\begin{minipage}{0.8\textwidth}
\caption{
Distribution of sampled similarity constraint values, where each horizontal sliver represents one independent observation.
}
\label{fig:sphere_distnplot}
\end{minipage}
\end{subfigure}
\end{minipage}

\caption{
Similarity constraint of tag-matching metrics.
}
\label{fig:sphere}

\end{center}
\end{figure*}
 
To characterize similarity constraint, we randomly sampled 5,000 target tags.
Then, for each target tag $R$ we randomly sampled tags until we found two secondarily-sampled tags $S_1$ and $S_2$ that were within a 0.01 match distance radius to the target.
Finally, we computed the match distance $d$ between the pair of secondarily-sampled tags.
Figure \ref{fig:dimensionality_measure} summarizes this process.

Figure \ref{fig:sphere_barplot} provides our estimate of the similarity constraint statistic for each metric, with error bars representing a 95\% confidence interval.
Figure \ref{fig:sphere_distnplot} shows the distribution of the similarity constraint statistic values among the 5,000 replicate samples in greater detail.

In a Euclidean space, similarity constraint corresponds to the average distance between points uniformly sampled from inside a ball (\textit{e.g.}, in two dimensions a circle, in three dimensions a sphere, \textit{etc.}).
In Euclidean space, this average distance increases with dimensionality.
For reference, in a one-dimensional Euclidean space similarity constraint would measure approximately 0.0067.
In a two dimensional Euclidean space, it would measure approximately  0.0091.
In 32 dimensions, it would measure 0.0137 \citep{dunbar1997average}.
So, in some sense, this similarity constraint metric can be interpreted as an indirect measure of dimensionality.
However, as we'll see in Section \ref{sec:detour_difference}, the hamming, hash, and streak metric impose a decidedly non-Euclidean geometry.

\subsubsection{Bidirectional Integer Metric}

For the bidirectional integer metric, we measured the similarity constraint statistic as 0.0068.
This falls in line with expectation: this metric is essentially identical to a one-dimensional Euclidean space.
As shown in Figure \ref{fig:sphere_distnplot}, the secondarily-sampled match distances are entirely bounded by the diameter of 0.02.
This metric not only exhibits tight similarity constraint in the mean case, but also permits no outliers to the similarity constraint.

\subsubsection{Integer Metric}

The integer metric exhibits much looser similarity constraint in the mean case.
We estimated this value as 0.5092.
However, this looser similarity constraint appears to be an artifact of averaging between two very tight constraints: a tight constraint to 0 in one case and a tight constraint to 1 in the other.
Figure \ref{fig:sphere_distnplot} confirms that all sampled match distances fall under one of these cases.
Because of the asymmetrical definition of the integer metric, half of pairs of similar scalar values will be in ascending order (resulting in a match distance close to 0) and half will be in descending order (resulting in wraparound search and a match distance close to 1).
The integer metric appears to allow for tags closely related to a third tag either very strongly match or very weakly match, but permits no intermediate outcomes.

\subsubsection{Hamming Metric}

The hamming metric exhibits a broader range of sampled similarity constraint values than the integer metrics.
We estimated mean similarity constraint as 0.1627, looser than the bidirectional integer metric.
As shown in Figure \ref{fig:sphere_distnplot}, many secondarily-sampled tag pairs are biased towards low match distances.
However, secondarily-sampled tag pairs that break this constraint are also not uncommon.
Among our 5000 trials, we observed distances between secondarily-sampled tags as high as 0.7499.

Why is our estimate of the hamming metric similarity constraint so much higher than the expected value of 0.0137 in a 32-dimensional Euclidean space?
This phenomenon appears to be due to the normalization process we applied to map raw match distances to a uniform distribution.
We also calculated this statistic for the raw hamming metric without normalization, increasing the radius of our sampling ball to 0.25.
(Only the exact target 32-bit tag itself falls within a sampling radius of 0.01.)
The \textit{a priori} expected distance between sampled points within a 32-dimensional ball with radius 0.25 is 0.3415.
Our estimate of similarity constraint for the raw hamming metric falls nearly in line with expectation at 0.3312.

\subsubsection{Streak Metric}

The streak metric exhibited the next-loosest similarity constraint statistic with a mean value sampled at 0.2813.
For this metric, we observed distances between secondarily-sampled tags as high as 0.9993.
The streak metric retains some geometric constraint in the mean case, but allows for outliers that strongly break similarity constraint.

\subsubsection{Hash Metric}

Like the unidirectional integer metric, the hash metric also exhibits a very loose similarity constraint of 0.5083 in the mean case.
However, unlike the integer metric, secondarily-sampled match distances are uniformly distributed between 0 and 1.
This is exactly as we would expect:
given any particular set of operands, a well-behaved hash function should yield a uniform distribution of hash results.
As expected, the hash metric exhibits no geometric structure.

\subsection{Dissimilarity Constraint} \label{sec:dissimilarityconstraint}

\begin{figure}

\begin{center}
\begin{subfigure}[b]{\linewidth}
\begin{minipage}{0.5\linewidth}

\includegraphics[width=0.75\linewidth]{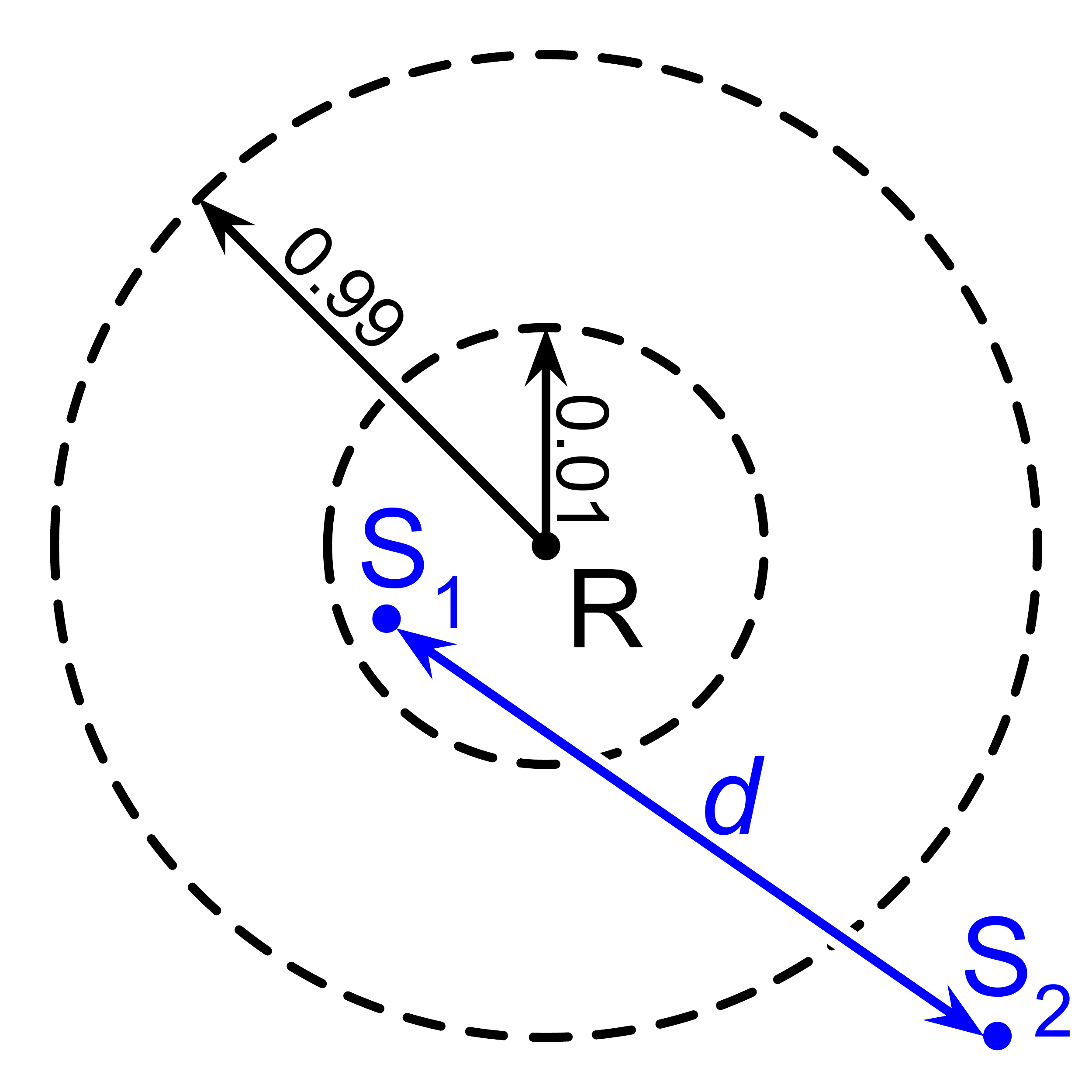}

\end{minipage}
\begin{minipage}{0.5\linewidth}
\caption{
A schematic depicting the process used to generate the dissimilarity statistic for each metric.
}
\label{fig:dissimilarity_statistic}
\end{minipage}

\end{subfigure}

\begin{subfigure}[b]{\linewidth}
\begin{minipage}{0.6\linewidth}
\includegraphics[width=\linewidth]{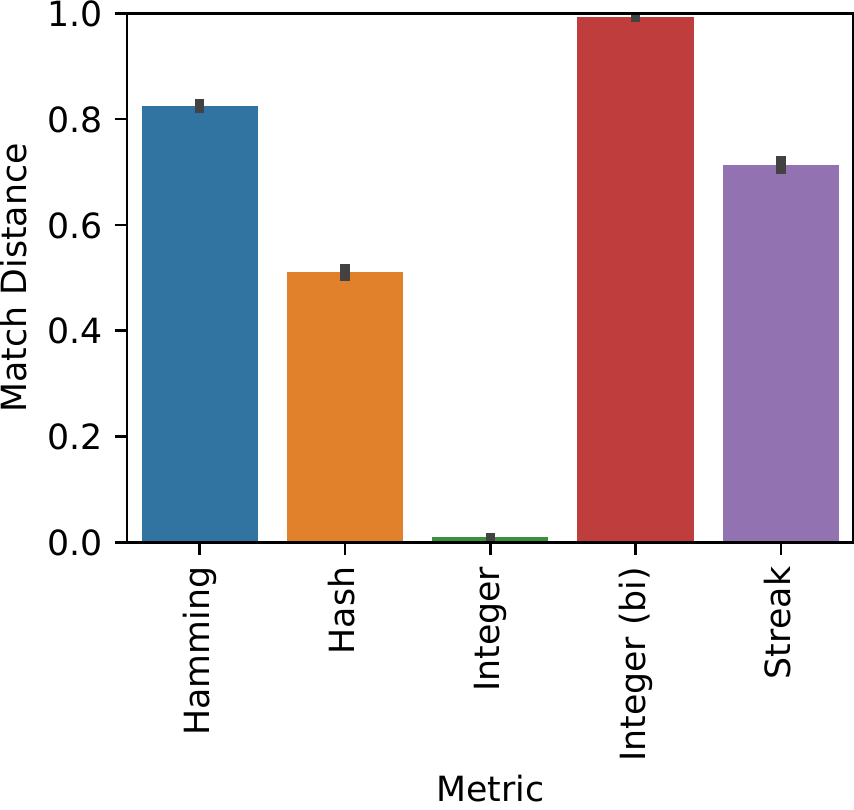}
\end{minipage}
\begin{minipage}{0.35\linewidth}
\caption{
Mean statistic values for each metric.
Error bars represent 95\% confidence intervals.
}
\label{fig:sphere_reverse_distnplot}
\end{minipage}
\end{subfigure}
s
\begin{subfigure}[b]{\columnwidth}
\centering
\includegraphics[width=\columnwidth]{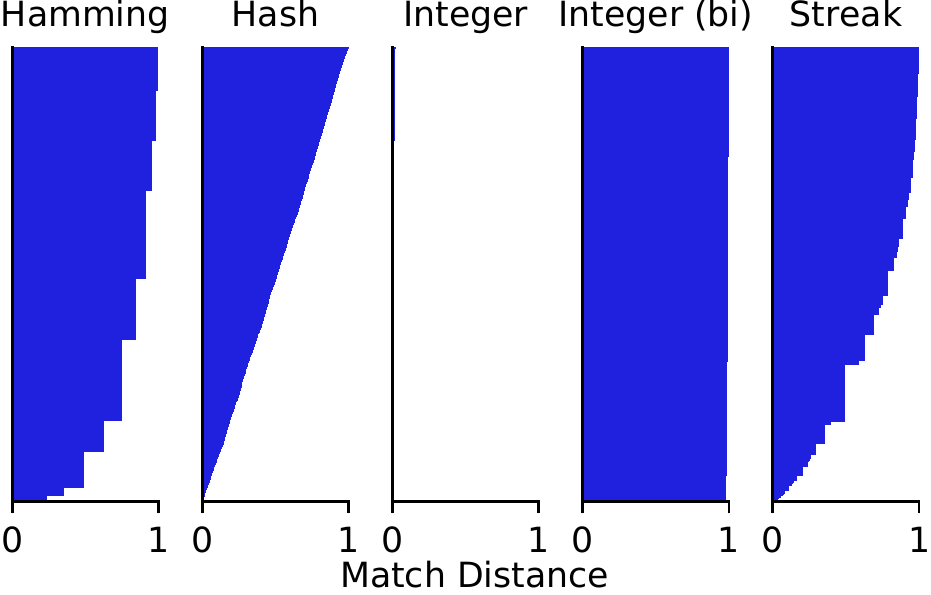}
\caption{
Statistic distribution, where each horizontal bar sliver represents one independent observation.
}
\label{fig:sphere_reverse_barplot}
\end{subfigure}

\caption{
Dissimilarity constraint of tag-matching metrics.
}
\label{fig:sphere_reverse}

\end{center}
\end{figure}

To characterize dissimilarity constraint, we randomly sampled 5,000 target tags.
Then, for each target tag $R$ we randomly sampled tags until we found a secondarily-sampled tag $S_1$ that was within a 0.01 match distance radius of $R$ and a secondarily-sampled tag $S_2$ that was outside a 0.99 match distance radius of the $R$.
Finally, we computed the match distance between $S_1$ and $S_2$.
Figure \ref{fig:dissimilarity_statistic} summarizes this process.

Figure \ref{fig:sphere_barplot} provides our estimate of the dissimilarity constraint statistic for each metric, with error bars representing a 95\% confidence interval.
Figure \ref{fig:sphere_distnplot} shows the distribution of the dissimilarity constraint statistic values among the 5,000 replicate samples in greater detail.

These results tell a similarity to similarity constraint.

\subsubsection{Hash Metric}
The hash metric exhibited no geometric structure --- $S_1$ and $S_2$ were uniformly likely to exhibit any match distance between 0 and 1.

\subsubsection{Streak Metric}

The streak metric exhibited some geometric structure in the mean case. We observed a mean secondarily-sampled distance 0.7127, significantly greater than the mean distance of 0.5 expected between arbitrarily-sampled tags.

However, outcomes that strongly broke geometric constraints also occurred.
We observed distances between secondarily-sampled tags as low as 0.0002.

\subsubsection{Hamming Metric}

The hamming metric exhibited stronger geometric structure in the mean case than the streak metric.
Mean secondarily-sampled distance was 0.8248.

This hamming metric also exhibited less extreme tail-end outcomes than the streak metric.
We observed match distances between the secondarily-sampled tags only as low as 0.2355.

\subsubsection{Bidirectional Integer Metric}

The bidirectional integer metric was highly constrained in both the mean and tail-end cases.
The smallest distance between secondarily-sampled tags observed was 0.9802.

\subsubsection{Integer Metric}

Again, the unidirectional integer metric exhibited a quirky result due to its noncommutative nature.
The mean distance between secondarily-sampled tags was 0.0100.
That is, instead of a bias against close matches as we would expect, secondarily-sampled tags were much closer together than expected under arbitrary sampling.
As shown in Figure \ref{fig:sphere_distnplot}, \textit{all} secondarily-sampled distances observed with this metric were extremely small.
So, although in the opposite way from what we would expect, match distances were still tightly constrained.

The mechanism behind this result stems from the metric's asymmetrical nature.
Under this metric, if you sample a tag that is close to a target it will be numerically slightly larger than the target.
Likewise, if you sample a tag that is very far from a target, it will be numerically slightly smaller than the target (due to wraparound).
Then, explaining this counterintuitive result, the distance from the slightly smaller to the slightly larger tag will be small.

\subsection{Detour Difference} \label{sec:detour_difference}

\begin{figure}
\begin{center}

\begin{minipage}{\linewidth}
\begin{subfigure}[b]{\linewidth}
\begin{minipage}{0.5\textwidth}
\begin{center}
\includegraphics[width=\linewidth,trim=2cm 5cm 2cm 5cm, clip]{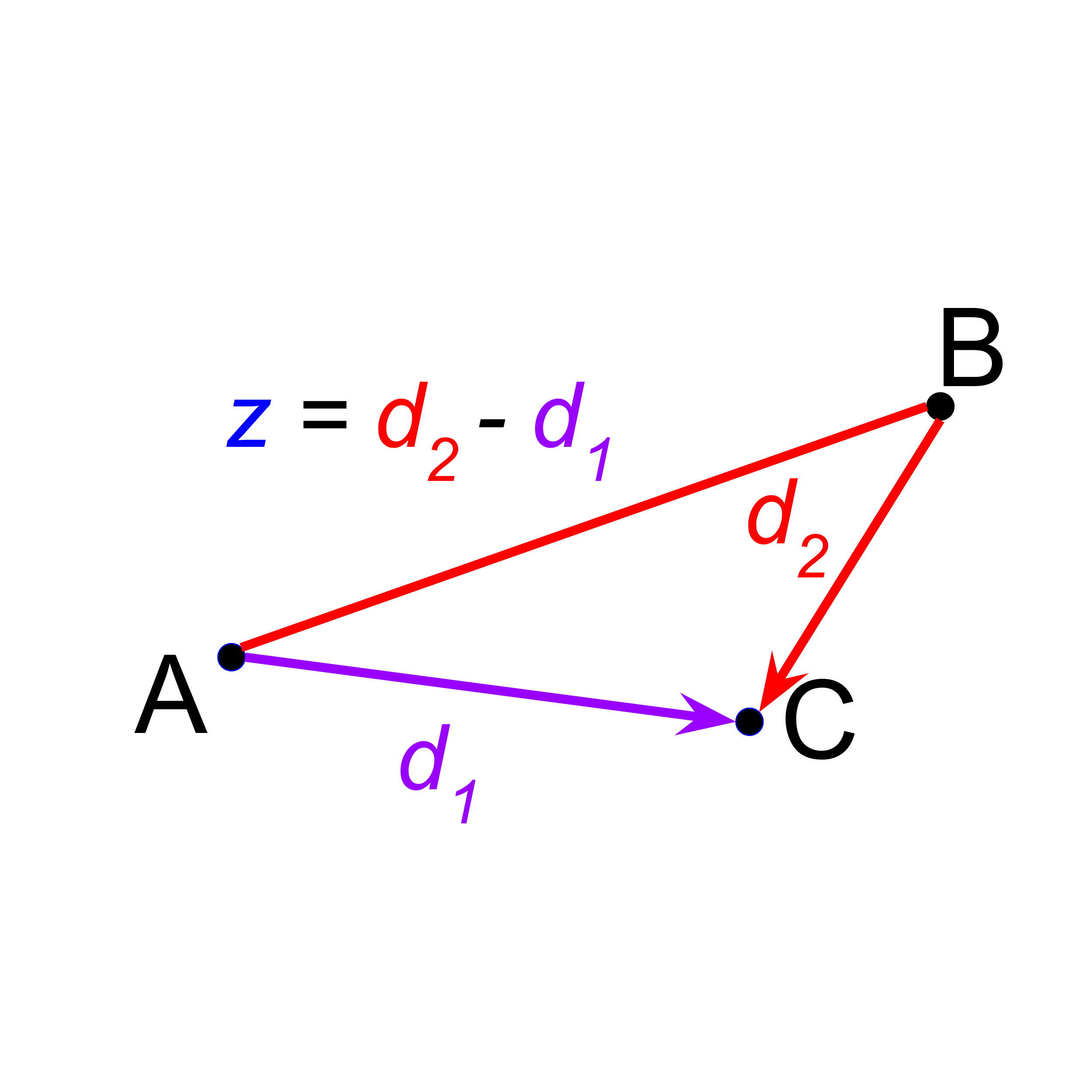}
\end{center}
\end{minipage}\begin{minipage}{0.5\textwidth}
\caption{
Sampling process used to evaluate detour difference, $z$.
} \label{fig:detour_difference_cartoon}
\end{minipage}
\end{subfigure}
\end{minipage}

\begin{minipage}{\linewidth}
\begin{subfigure}[b]{\linewidth}
\includegraphics[width=\linewidth]{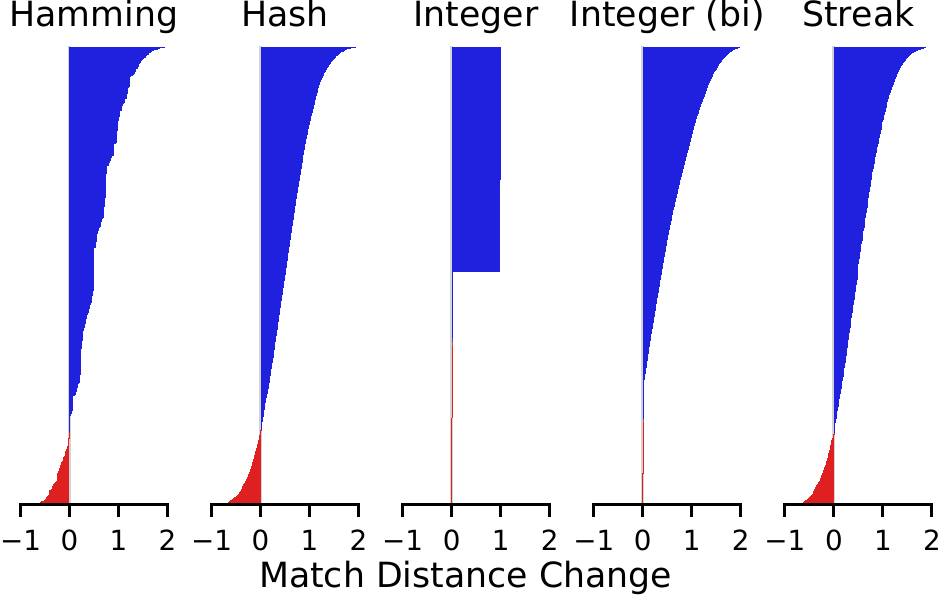}
\caption{
Distributions of detour distance difference for triplets of randomly sampled tags.
Each bar sliver represents an independently sampled observation.
A positive value (colored blue) indicates that total distance increased with the addition of an intermediate stop.
A value of exactly 0 indicates an intermediate stop had no effect on total distance.
A negative value (colored red) indicates violation of the triangle inequality: taking an intermediate stop reduced the total distance travelled.
} \label{fig:detour_difference_distribution}

\end{subfigure}
\end{minipage}

\caption{
Detour difference of tag-matching metrics.
}
\label{fig:detour_difference}

\end{center}
\end{figure}
 
Similarity constraint and dissimilarity constraint quantify the geometric constraint imposed under preexisting strong matching and strong mismatching, respectively.
To complement these measures, we set out to characterize the regularity, in a loose sense, of each space more broadly.
This led us to our ``detour difference'' measure, which quantifies how tag matching spaces respect the triangle inequality.

Intuitively, detour difference is a measure of how adding a randomly-chosen waypoint affects total distance between a pre-existing start and end.
Under the triangle inequality, the direct route is always shortest.
So, if the triangle inequality is respected, detour difference should always be non-negative.

To measure detour difference, we uniformly sampled 5,000 triplets of tags $A$, $B$, and $C$.
Then, for each metric $m$ we calculated the $m(A, B) + m(B, C) - m(A, C)$.
Figure \ref{fig:detour_difference_cartoon} provides a schematic of this process.

Figure \ref{fig:detour_difference_distribution} plots the distribution of the detour difference statistic for each metric.
The hamming, hash, and streak metrics show evidence of ``shortcuts'' that violate the triangle inequality.
\footnote{
The raw hamming metric does respect the triangle inequality, so presumably this result is due to the normalization.
}
Surprisingly, given results from the similarity and dissimilarity constraint measures, the distributions of detour difference for these three metrics appear very similar.
This suggests that geometric differences between these metrics are specially accentuated in contexts of preexisting strong matching and mismatching constraint.
 
\section{Variational Analysis} \label{sec:variational}

In Section \ref{sec:geometric} we investigated how existing tag-match relationships to a common tag influenced the distribution of match distances.
This section, in contrast, focuses on how individual bit-flip mutations --- and cumulative sequences of bit-flip mutations --- affect a single tag-matching relationship.

In Section \ref{sec:single_step}, we report two single-step mutational analyses: one that examines the local mutational neighborhoods of loosely-affiliated and a second that examines the local mutational neighborhoods of tightly-affiliated tag pairs.
In Section \ref{sec:mutational_walks}, we perform mutational walk analysis to survey the broader mutational landscape.

\subsection{Single-Step Mutations} \label{sec:single_step}

\begin{figure}
\begin{center}

\includegraphics[width=\columnwidth]{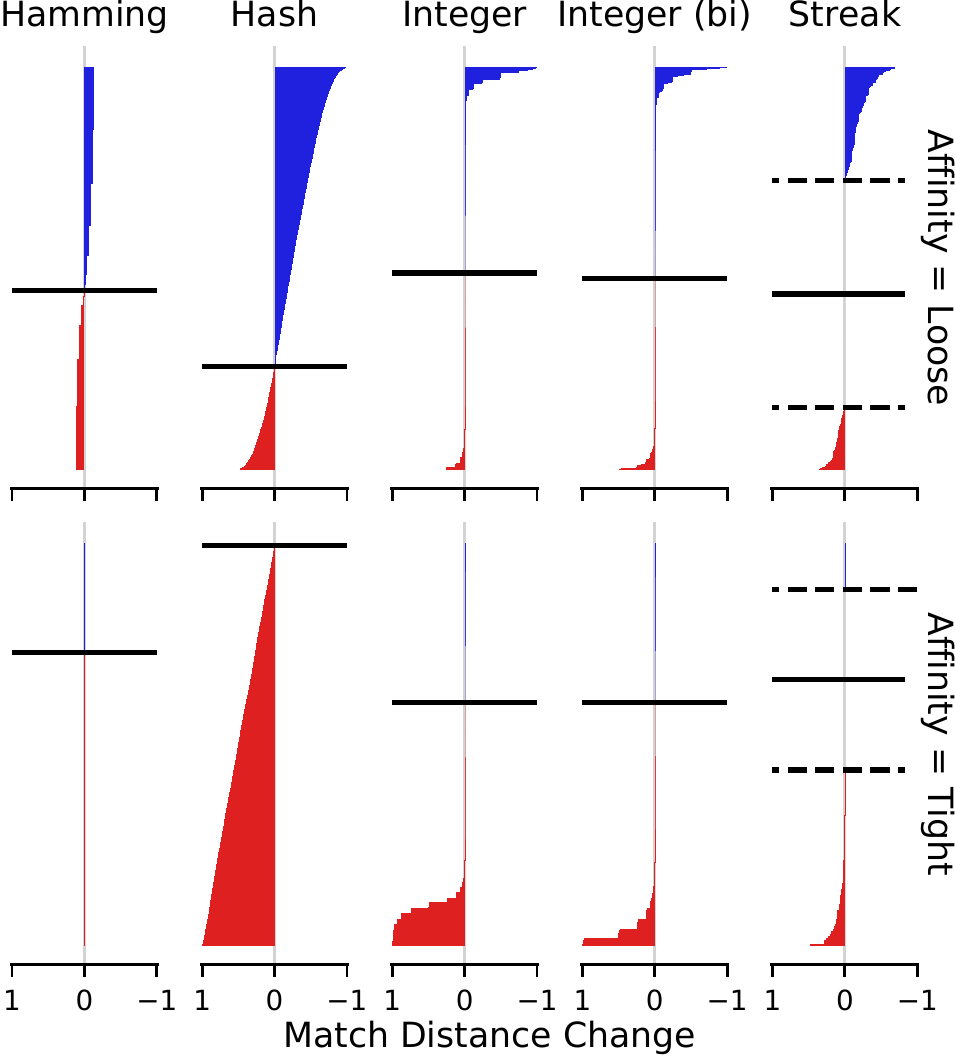}
\caption{
Distributions of mutation effects on match distance for loosely matched (pre-mutation match distance $> 0.5$) and tightly matched (pre-mutation match distance $< 0.01$) tag pairs.
Each bar sliver represents a single independently sampled mutation on an independently sampled tag pair.
Mutations that increase affinity are colored blue and mutations that decrease affinity are colored red.
Solid lines indicate the median between mutations that increase match distance and mutations that decrease match distance.
Dashed lines demarcate the boundaries between non-neutral and perfectly-neutral mutations.
}
\label{fig:mutational_step}

\end{center}
\end{figure}
 
We performed single-step mutational analyses to characterize the local mutational neighborhoods induced by each tag-matching metric.

To measure this effect of mutation on loosely-affiliated tag pairs, we
\begin{itemize}
    \item randomly sampled a target tag $R$,
    \item randomly sampled candidate tags until we found a second tag $S$ with a match distance $> 0.5$,
    \item recorded match distance $d$ between $R$ and $S$,
    \item applied a one-bit mutation to the secondary tag $S$, yielding a mutated variant $S'$, 
    \item measured the match distance $d$ between $R$ and $S'$,
    \item and then calculated change in match distance under mutation $p = d' - d$.
\end{itemize}
We repeated this procedure to generate 5,000 samples.

The top panel of Figure \ref{fig:mutational_step} visualizes the distribution of match distance change under mutation of loosely-affiliated tags.
A negative mutational perturbation $p$ indicates a decrease in match distance and, therefore, an increase in match quality (colored blue in Figure \ref{fig:mutational_step}).
A positive mutational perturbation $p$ indicates an increase in match distance and, therefore, an decrease in match quality (colored red in Figure \ref{fig:mutational_step}).

We measured the distribution of mutational perturbations on tightly-matched tag pairs similarly, except we uniformly sampled until we found a second tag $S$ with match distance $< 0.01$.
The bottom panel of Figure \ref{fig:mutational_step}, color coded identically to the top panel, visualizes the distribution of match distance change under mutation of loosely-affiliated tags.
This distribution reflects the effects of one-step mutations on tags with pre-existing affinity.

\subsubsection{Integer and Bidirectional Integer Metrics}

For both tightly- and loosely-affiliated tag pairs under the integer and bidirectional integer metrics, most mutations caused very small changes in match distance.
These mutations toggle least-significant bits of the tag's integer representation.
However, under these metrics, a small fraction of mutations affecting more-significant bits of the integer representation have a much stronger effect.
Single-step mutations occasionally occur that strongly couple loosely-affiliated tag pairs or strongly decouple tightly-affiliated tag pairs.
In particular, the unidirectional integer metric appears to exhibit more frequent strong decoupling mutations than the bidirectional integer metric, presumably due to its non-commutative quirks.

\subsubsection{Streak Metric}

The streak metric exhibits a large fraction of perfectly neutral outcomes under mutation.
These perfectly-neutral mutations presumably affect regions of the bitstring neither involved in the longest-matching streak nor in the longest-mismatching streak.
The streak metric exhibits a thicker tail of mutational magnitude for mutations that couple loosely-affiliated tags than the integer metrics.
In addition, the most extreme mutational outcomes that couple loosely-affiliated tags appear to be of a comparable magnitude to those under the integer metrics.
Mechanistically, this might be due to mutations that disrupt longest-mismatching streaks.
However, one-step mutations that decouple tightly-affiliated tags do not appear as potent.
This might be because achieving a very poor match requires both increasing longest-mismatching streak length and decreasing longest-matching streak length.

\subsubsection{Hamming Metric}

The hamming metric exhibits a generally uniform magnitude of match-distance changes under mutation.
High-magnitude one-step mutations do not occur under this metric.
(Without normalizing match distance to a uniform distribution for randomly-sampled tags, all hamming metric mutations would be of exactly the same magnitude, either increasing or decreasing the count of matching bits by 1.)

\subsubsection{Hash Metric}

The hash metric exhibits the thickest tails of mutational magnitude of all metrics.
Extreme-effect one-step mutations are plentiful under this metric.
Interestingly, compared to other metrics, the hash metric exhibits a greater fraction of mutations that decouple tightly-affiliated tags and a greater fraction of mutations that couple loosely-affiliated tags.
This result can be attributed to the hash metric's lack of geometric structure.
Because all one-step mutations uniformly sample a new match distance, 99.5\% of one-step mutations on tightly-affiliated tags will result in a looser coupling.
Similarly, approximately 75\% of one-step mutations on loosely-affiliated tags will result in a tighter coupling.

\subsection{Mutational Walks} \label{sec:mutational_walks}

\begin{figure}
\begin{center}

\includegraphics[width=\textwidth]{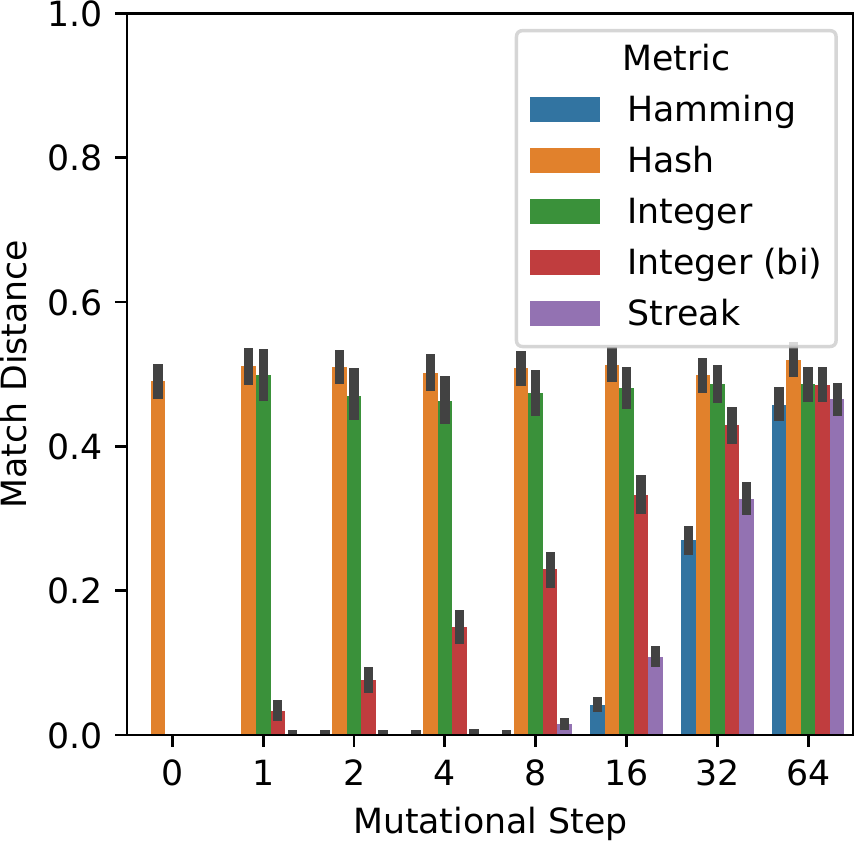}
\caption{
Match distance along mutational walks from initially identical tags.
Error bars represent 95\% confidence intervals.
Note lotharithmic scale on the $x$ axis.
}
\label{fig:mutational_walk_barplot}

\end{center}
\end{figure}
 
We performed single-step mutational analyses to characterize the broader mutational landscapes induced by each tag-matching metric.

To conduct a mutational walk, we 
\begin{itemize}
    \item randomly generated a starting tag,
    \item then sequentially applied 65 randomly-chosen one-step bit flip mutations (with back mutation allowed),
    \item while measuring match distance to the original starting tag at each step along the walk.    
\end{itemize}
We analyzed 1,000 replicate mutational walks for each metric.

Figure \ref{fig:mutational_walk_barplot} compares how match distance increases along mutational walks for each tag-matching metric.

\subsubsection{Hash Metric}

Due to the hash metric's lack of geometric structure the hash metric, bitwise equivalent tags do not exhibit low match distance.
So, as expected, throughout the entire mutational walk this metric maintains a constant mean match distance of 0.5.

\subsubsection{Integer Metric and Bidirectional Integer Metric}

The unsigned representation of the integer and bidirectional integer metrics (Section \ref{sec:integer} induces an exponential distribution of mutational effect across bits: mutating the most-significant bit has $2^n$-larger effect on the bitstring's integer value compared to the least-significant bit.
Such large-effect mutations provides a reasonable explanation for the bidirectional integer metric's rapid increase in match distance under mutation relative to the hamming and streak metrics. 
The integer metric experiences even more rapid dilation of match distance under mutation.
Under this metric, half of first mutational steps cause a wraparound effect, immediately spiking the average match distance to 0.5.
Supplementary Figure \ref{fig:mutational_walk_lineplot} shows match distance variance decreasing as the mutational walk proceeds away from match distances biased to 0 or 1.

\subsubsection{Hamming Metric}

The hamming metric's match distance diffuses upward slowest.
The hamming metric's mutational walk is significantly slower to diverge than the streak metric's at 16 and 32 steps (non-overlapping 95\% CI).
It is significantly slower to diverge than the integer metrics and the hash metric between steps 1 and 32, as well (non-overlapping 95\% CI).

\subsubsection{Streak Metric}

The streak metric diffuses away from zero match distance second-slowest, trailed only by the hamming metric.

Interestingly, this result contradicts Downing's presentation of the streak metric in \citep{downing2015intelligence}, in which he suggests that the streak metric exhibits greater robustness because its match distance diverges more slowly under a mutational walk.
This discrepancy presumably arises due to our normalization to ensure a uniform distribution of raw match scores between 0 and 1.
We believe that our result under normalization is more representative because match distance corresponds to the probability that arbitrary tags would match more strongly by chance --- which directly relates to how effectively a operand tag competes to be the ``best'' match for a query.

To compare mutational landscapes between the streak and integer metrics under more realistic circumstances (i.e., where tags do not begin arbitrarily perfectly-matched), we performed a secondary mutational walk experiment.
This experiment was conducted exactly as before, except instead of starting with exactly-identical tags it started with a pair of tags that was randomly sampled for match distance $<0.01$.

As shown in Supplementary Figure \ref{fig:mutational_walk_sampled_start_barplot}, this experiment confirmed greater robustness of the hamming metric under mutation.
The streak metric's match distance was significantly greater than the hamming metric between mutational steps 2 and 16 (non-overlapping 95\% CI).
Our result remained consistent when replicating the experiment with 64-bit tags (Supplementary Figure \ref{fig:mutational_walk_sampled_start_64_barplot}).

\section{Evolutionary Analysis} \label{sec:evolutionary}

Sections \ref{sec:geometric} and \ref{sec:variational} reported how each tag-matching metric induced constraints on tag-match affinities and the distribution of mutational outcomes.
We now move on to investigate whether --- and how ---  these geometric and variational properties affect evolution of tag-mediated connectivity in various scenarios.

We begin with a toy problem, presented in Section \ref{sec:graph-matching}, which allowed us to systematically vary the level of network constraint selected for.
That is, these experiments compared scenarios where individual tags needed to ensure simultaneously tight affinity with several other tags (more constrained) and where individual tags only needed to ensure tight affinity with one other tag (less constrained).
In this problem, we define a target connection topology between tagged queries and operands then select for sets of tags that exhibit high-affinity pairings between connected topology elements.

In order to investigate potential consequences of tag-matching metrics in a more generalized, complex domain, we evolved full-fledged SignalGP programs that mediate module activation via tag matching.

The SignalGP genetic programming representation employs tag-based referencing to facilitate event-driven program execution \citep{lalejini2018evolving}.
In SignalGP, programs are segmented into modules (functions) that may be automatically triggered by exogenously- or endogenously-generated signals.
Tags specify the relationship between signals and signal-handlers (program modules), triggering the module with the closest matching tag to run its linear sequence of instructions.

The SignalGP instruction set, in addition to including traditional GP operations, allows programs to generate arbitrarily-tagged internal signals and broadcast arbitrarily-tagged external signals, and otherwise work in a tag-based context.
SignalGP also supports genetic regulation with promoter and repressor instructions that, when executed, allow programs to adjust how well subsequent signals match with a target function (specified with tag-based referencing) \citep{lalejini2021tag}.
See \cite{lalejini2018evolving} for a more detailed description of SignalGP.

To ensure a broad survey of tag-matching functionality, we performed experiments with a complementary pair of SignalGP problems:
\begin{itemize}
    \item the \textit{Changing-signal Task} (Section \ref{sec:changing-signal}), which is known to select for sparse tag interactions (i.e., low constraint), and
    \item the \textit{Directional-signal Task} (Section \ref{sec:directional-signal}), which is known to select for more dense tag interactions (i.e., high constraint).
\end{itemize}

\subsection{Graph-matching Task} \label{sec:graph-matching}

\begin{figure}
\begin{center}

\begin{minipage}{0.05\textwidth}
~
\end{minipage}\begin{minipage}{0.95\textwidth}
\begin{minipage}{0.05\textwidth}
~
\end{minipage}\begin{minipage}{0.95\textwidth}
\centering
\large
\textbf{Mean Degree}
\end{minipage}
\begin{minipage}{0.05\textwidth}
~
\end{minipage}\begin{minipage}{0.95\linewidth}
\begin{minipage}{0.5\textwidth}
\centering
\large
1
\end{minipage}\begin{minipage}{0.5\textwidth}
\centering
\large
2
\end{minipage}
\end{minipage}
\end{minipage}\\
\vspace{2ex}

\begin{minipage}{0.05\textwidth}
\large
\rotatebox[origin=c]{90}{\textbf{Structure}}
\end{minipage}\begin{minipage}{0.95\textwidth}
\begin{minipage}{0.05\linewidth}
\large
\rotatebox[origin=c]{90}{Irregular}
\end{minipage}\begin{minipage}{0.95\linewidth}
\begin{subfigure}[b]{0.5\textwidth}
\centering
\includegraphics[width=\textwidth]{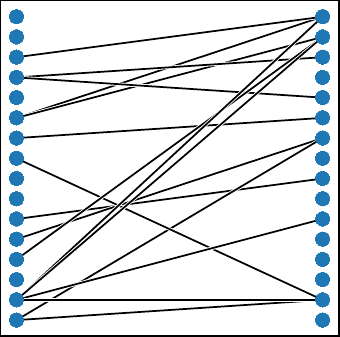}\caption{
Irregular w/ mean degree 1
}
\label{fig:irregular_1}
\end{subfigure}
\begin{subfigure}[b]{0.5\textwidth}
\centering
\includegraphics[width=\textwidth]{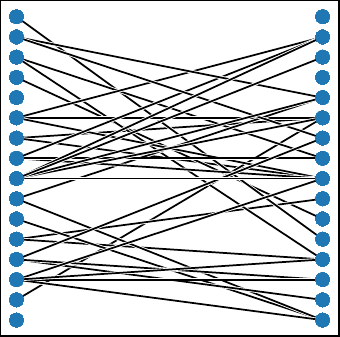}\caption{
Irregular w/ mean degree 2
}
\label{fig:irregular_2}
\label{fig:irregular_degree_2}
\end{subfigure}

\end{minipage}

\vspace{2ex}

\begin{minipage}{\textwidth}

\begin{minipage}{0.05\linewidth}
\large
\rotatebox[origin=c]{90}{Regular}
\end{minipage}\begin{minipage}{0.95\linewidth}
\begin{subfigure}[b]{0.5\textwidth}
\centering
\includegraphics[width=\textwidth]{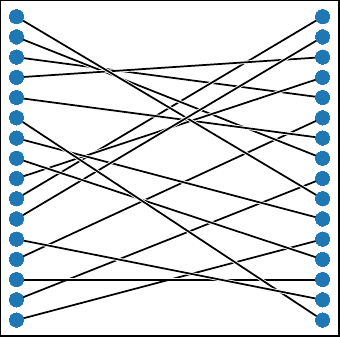}\caption{
Regular w/ mean degree 1
}
\label{fig:regular_degree_1}
\label{fig:regular_1}
\end{subfigure}
\begin{subfigure}[b]{0.5\textwidth}
\centering
\includegraphics[width=\textwidth]{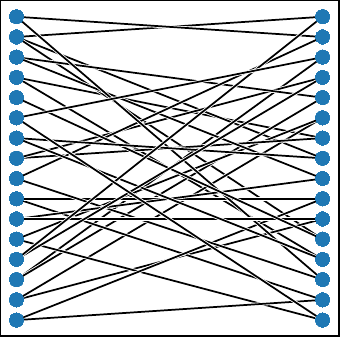}\caption{
Regular w/ mean degree 2
}
\label{fig:regular_degree_2}
\label{fig:regular_2}
\end{subfigure}
\end{minipage}
\end{minipage}
\end{minipage}

\caption{
Example target graph layouts used in 32-node graph-matching evolutionary experiments.
Blue dots represent tagged nodes.
Black lines represent selected-for tight affinity relationships.
Layouts differ in total number of selected-for affinities (``mean degree'') and whether selected-for affinities were evenly or randomly distributed between nodes (``structure'').
}
\label{fig:graph_layouts}

\end{center}
\end{figure}
 \begin{figure*}

\includegraphics[width=\linewidth]{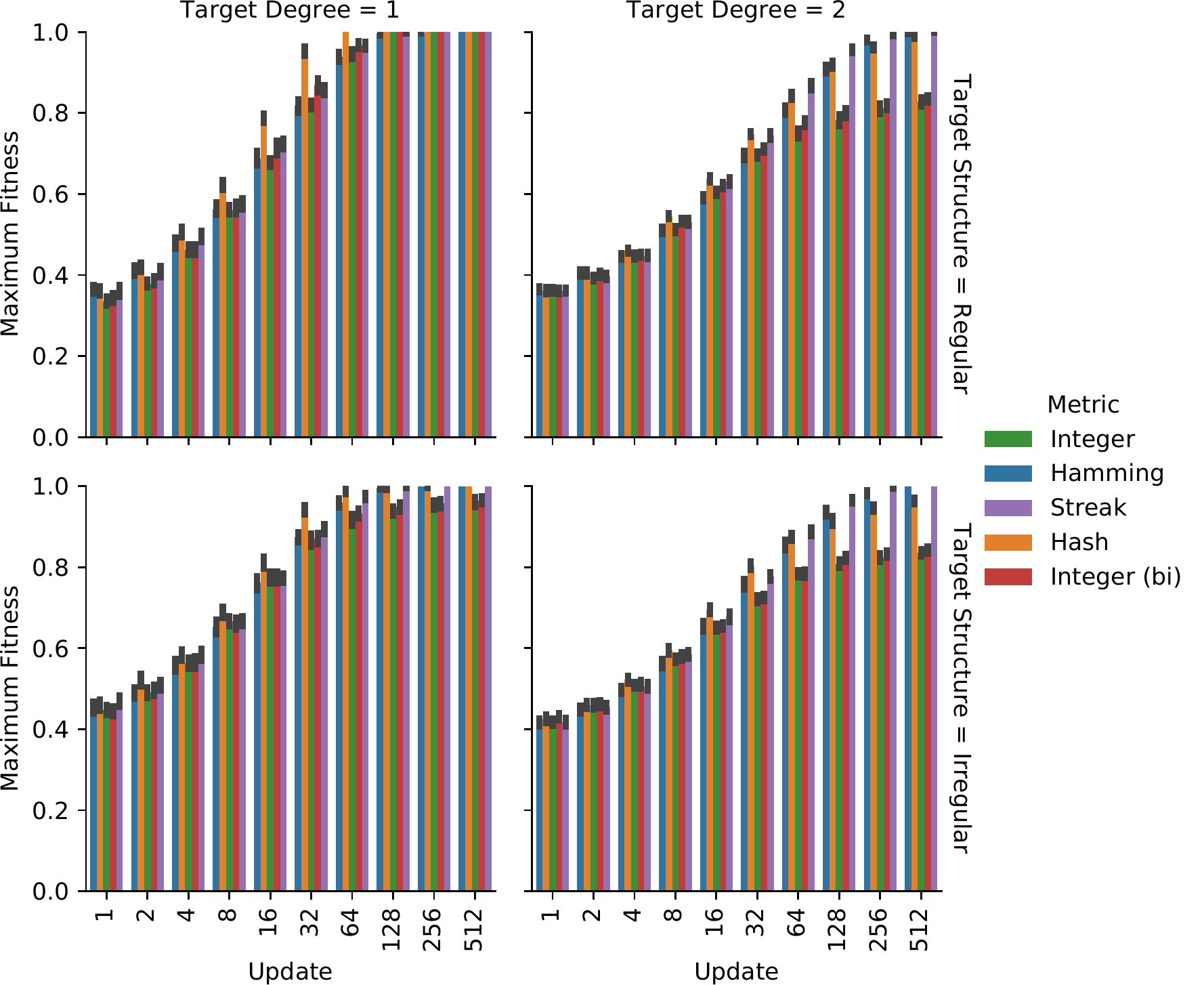}

\caption{
Trajectories of adaptive evolution for each tag-matching metric on the 64-node graph-matching task.
Maximum fitness represents the best fitness value for any individual within a population.
Here report using each metric's best-performing per-bit mutation rate.
(See Supplementary Figure \ref{fig:evolve_mutsweep} for survey showing how mutation rate affects adaptive evolution under each metric.)
Note log-scale x-axes.
Shaded area represents bootstrapped 95\% confidence intervals across 20 replicate observations.
}
\label{fig:evolve_bests}
\end{figure*}
 
In this evolutionary experiment, we evolved genomes consisting of 32 bitstring tags to establish a pattern of connectivity exactly mirroring that of a randomly-generated target bipartite graph.
Each bitstring tag in a genome corresponded to a node in the target graph.
Figure \ref{fig:graph_layouts} shows example target graph layouts.

Target graphs were evenly partitioned between queries and operands.
To evaluate the fitness of a genome, we harvested its operand tags placed them into a tag-matching data structure.
This data structure allowed us to determine the best-matching operands for each query tag.
We determined best matches as the operand tags with lowest match distance to that query.
For each query tag, we recorded as many best-match results as the number of outgoing edges on the corresponding node in the target graph.
We assessed fitness as the fraction of best-match tag pairs that correctly corresponded to edges in the target graph.

We controlled the degree of tag-matching constraint imposed by the target graph by manipulating:
\begin{enumerate}
  \item \textit{mean degree} --- the number of edges between queries and operands, and
  \item \textit{structure} --- whether edges were assigned evenly such that all nodes had identical degree (regular structure) or were assigned at random, likely causing some nodes to have high degree (irregular structure).
\end{enumerate}
We tested target graphs mean degree 1 and 2 and both regular and irregular construction.

Irregular, degree 2 graphs imposed the most tag-matching constraint.
High-degree nodes in these graphs were exceptionally constrained by many simultaneous connection criteria.
Figure \ref{fig:irregular_degree_2} shows an example irregular, degree 2 graph.

Regular, degree 1 graphs imposed the least tag-matching constraint.
Figure \ref{fig:regular_degree_1} shows an example regular, degree 1 graph.

For each target graph configuration, we surveyed each metric's performance over ten per-bit mutation rates ranging from 0.75 expected bit mutations per genome to 16.0 expected bit mutations per genome.
For each combination of metric and target graph configuration, we report results from the most favorable mutation rate (as defined by sum population-maximum fitness across updates).
\footnote{
Supplementary Figure \ref{fig:evolve_mutsweep} shows the rate each metric's rate of adaptive evolution across surveyed mutation rates for each target graph configurations.
All treatments' optimal mutation rates fall within the range of mutation rates surveyed, except for the hash metric on the regular target graphs.
In this case, peak performance was observed on the lowest sampled mutation rate.
}

We ran 100 replicate 512-generation evolutionary runs for each mutation rate/target graph/tag-matching metric combination.
These runs had a well-mixed population of size 500 and used tournament selection with tournament size 7.
Figure \ref{fig:evolve_bests} plots population-maximum fitness over the course of these evolutionary runs.
We performed the same evolutionary experiment with larger 64-node target graphs and observed qualitatively similar results (Supplementary Figures \ref{fig:evolve_bests64} and \ref{fig:evolve_big_mutsweep}).

\subsubsection{Hash Metric}

Surprisingly, the hash metric enables faster adaptive evolution than all other metrics on the least-constrained target graph (Figure \ref{fig:evolve_bests}; non-overlapping 95\% CI).
On more-constrained target graphs with mean degree 2, the hash metric's advantage in rapid adaptive evolution disappears.
In fact, on the most-constrained target graph (irregular structure with mean degree 2) the hash metric yields significantly lower-quality solutions at the end of evolutionary runs than the streak and hamming metrics (Figure \ref{fig:evolve_bests}; non-overlapping 95\% CI).

\subsubsection{Integer Metrics}

The integer and bidirectional integer metrics successfully match the least-constrained target graph (regular structure with mean degree 1) but yield lower-quality solutions than other metrics on more constrained target graphs (Figure \ref{fig:evolve_bests}; non-overlapping 95\% CI).

\subsubsection{Streak Metric}

The streak metric facilitates slightly faster adaptive evolution than the hamming metric, especially on mean degree 2 regularly configured target graphs (Figure \ref{fig:evolve_bests}; non-overlapping 95\% CI).

\subsection{Changing-signal Task} \label{sec:changing-signal}

\begin{figure}
\centering
\includegraphics[width=0.75\linewidth]{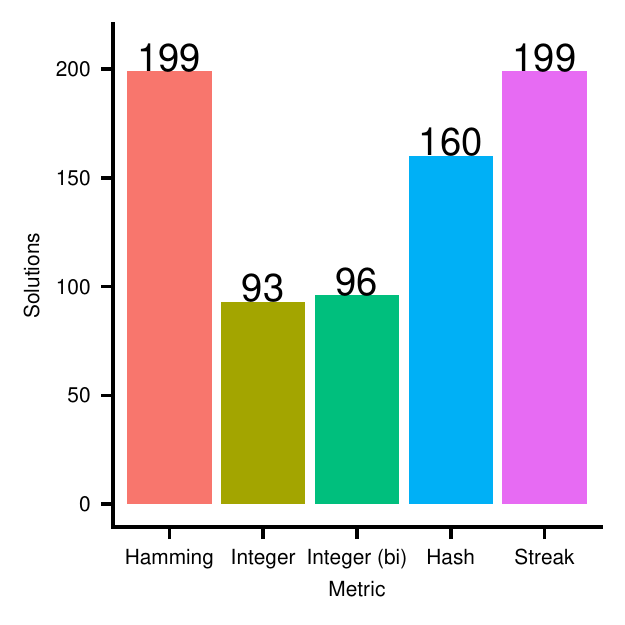}

\caption{
Evolutionary performance of tag-matching metrics on the changing signals task.
Shows the numbers of replicates out of 200 that produced a complete task solution to the changing-signal and directional-signal task respectively.
Results for each metrics' best-performing mutation rate are reported.
}

\label{fig:cst-sols}

\end{figure} 
The changing-signal task requires SignalGP programs to express a certain, distinct response to each of $K$ environmental signals.
Environmental signals correspond to a unique tagged event.
Programs express a response by executing one of $K$ response instructions.
Successful programs can ``hardcode'' each response to the appropriate environmental signal by ensuring that each environmental signal's tag best matches the function containing its correct response.
Thus, in this experiment SignalGP module tags are minimally constrained --- each needs to only match with a single environmental signal.

During evaluation, we afford programs 64 virtual CPU cycles to express the appropriate response after receiving a signal.
Once a program expresses a response or the allotted time expires, we reset the program's virtual hardware (resetting all executing threads and thread-local memory), and the environment produces the next signal.
Evaluation continues until the program correctly responds to each of the $K$ environmental signals or until the program expresses an incorrect response.
During each evaluation, programs experience environmental signals in a random order; thus, the correct \textit{order} of responses will vary and cannot be hardcoded.

For each tag-matching metric, we evolved 200 replicate populations (each with a unique random number seed) of 500 asexually reproducing programs in an eight-signal environment ($K=8$) for 100 generations.
We identified the most performant per-bit tag mutation rates (from a range of possible mutation rates) for each metric on the changing-signal task:
\begin{itemize}
    \item 0.01 for the hamming and streak metrics,
    \item 0.002 for the hash metric, and
    \item 0.02 for the integer and bidirectional integer metrics.
\end{itemize}
Aside from tag mutation rate, the overall configuration used for each metric was identical.

We limited tag variation in offspring to tag mutation operators (bit flips) by initializing populations with a common ancestor program in which all tags were identical and by disallowing mutations that would insert instructions with random tags.
Supplemental Section \ref{sec:gpsupplement} gives the full configuration details for this experiment, including a guide for replication.

Figure \ref{fig:cst-sols} gives the number of replicates that produced a successful SignalGP program (i.e., capable of achieving maximum fitness) for each tag-matching metric on the changing-signal task.
We compared the number of successful replicates across metrics using a pairwise Fisher's exact test with a Holm correction for multiple comparisons.

\subsubsection{Hamming and Streak Metrics}

The hamming and streak metrics performed significantly better than all other metrics ($p < 5\times10^{-11}$); however, there was no significant difference in performance between the hamming and streak metrics.
To assess whether the streak metric produced solutions in fewer generations than the hamming metric, we ran 200 new replicates of each condition until 100 replicates produced a solution and recorded the number of generations that elapsed (Supplementary Figure \ref{fig:cst-times}).
We found no difference in generations elapsed between the hamming and streak metrics.

\subsubsection{Hash Metric}

The hash metric significantly outperformed both integer metrics ($p < 4\times10^{-10}$).

We suspect that the hash metric performed well because it maximizes generation of phenotypic variation (i.e., signal-function relationships).
Even a single bit flip in a tag is likely to completely re-order which other tags it best matches with.
The capacity to quickly generate large amounts of phenotypic variation allows evolution to explore large swaths of the fitness landscape from generation to generation, which is particularly useful in this low-constraint problem.
However, as evidenced by better performance of the hamming and streak metrics, this capacity to generate phenotypic variation trades off with tag-matching robustness --- under this metric, a single bit mutation may also scramble established relationships with other tags.

\subsubsection{Integer Metrics}

Among surveyed tag-match metrics, the integer metrics performed worst.
We observed no adaptive difference  between the integer and bidirectional integer metrics.

\subsection{Directional-signal Task} \label{sec:directional-signal}

\begin{figure}

\begin{subfigure}[b]{\linewidth}
\centering
\includegraphics[width=0.75\linewidth]{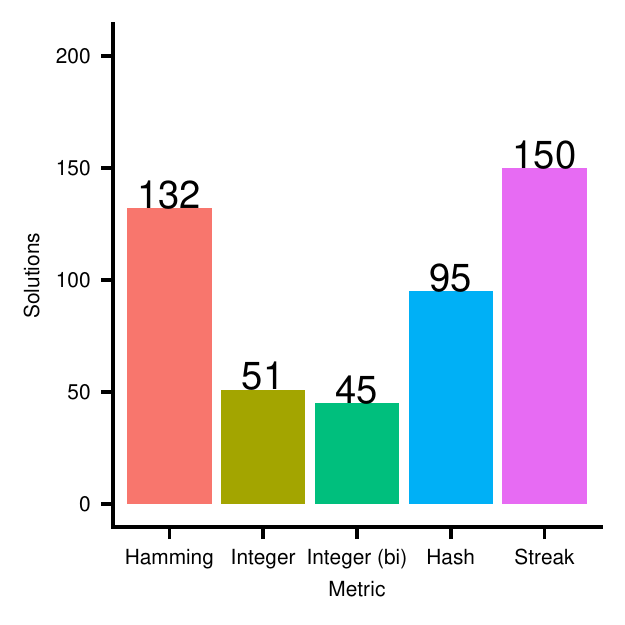}\caption{
Numbers of replicates out of 200 that produced a complete solution to the directional-signal task.}
\label{fig:dst-sols}
\end{subfigure}
\begin{subfigure}[b]{\linewidth}
\centering
\includegraphics[width=0.75\textwidth]{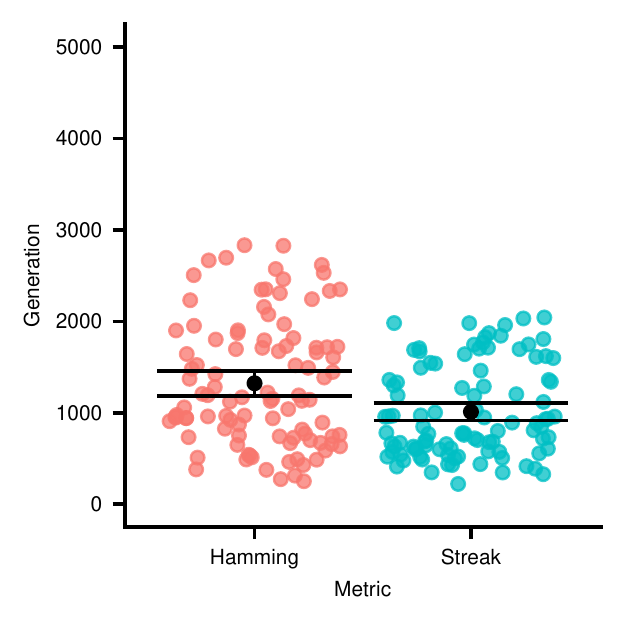}\caption{
Generations to solution for the first 100 replicates out of 200 to produce a complete solution to the directional-signal task.
Error bars indicate bootstrapped 95\% confidence intervals.
}
\label{fig:dst-times}
\end{subfigure}

\label{fig:gp_results}

\caption{
Evolutionary performance of tag-matching metrics on the directional signals.
All show each metrics' best-performing mutation rate.
}

\label{fig:evo_directional_signal}
\end{figure} 
As in the changing-signal task, the directional-signal task requires that programs respond to a sequence of environmental cues.
In the directional-signal task, however, the correct response to signal depends on the history previously experienced signals.
In the directional-signal task, there are two possible environmental signals --- a ``forward signal'' and a ``backward signal'' (each with a distinct tag) ---  and a cycle of four possible responses.
If a program receives a forward-signal, it should express the next response in the cycle.
If the program receives, a backward-signal, it should express the previous response in the cycle.
For example, if response three is currently required, a subsequent forward signal indicates that response four is required next, while a backward signal would instead indicate that response-two is required next.
Because the appropriate response to both the backward and forward signals change over time, successful programs must regulate which functions these signals trigger (rather than hardcode each response to a particular signal).

SignalGP module tags are more constrained than in the changing-signal task, potentially needing to match to queries by genetic regulation instructions in addition to \textit{several} tagged events (e.g., environmental signals or internally-generated signals) depending on internal regulatory state.
Indeed, in other work, we have observed that the directional signal task yields significantly more interconnected regulatory networks than the changing signal task \citep{Lalejini_Moreno_Ofria_2020}.

We evaluate programs on all possible four-signal sequences of forward and backward signals (sixteen total).
For each program, we evaluate each sequence of signals independently, and a program's fitness is equal to its aggregate performance.
Otherwise, evaluation on a single sequence of signals mirrors that of the changing signal task.

We used an identical experimental design for the directional-signal task as in the changing signal task.
However, we evolved programs for 5,000 generations (instead of 100) and re-parameterized each metric's tag mutation rate:
\begin{itemize}
\item 0.001 for the hamming and hash metrics,
\item 0.002 for the integer and streak metrics, and
\item 0.0001 for the bidirectional integer metric.
\end{itemize}
Full configuration details for this experiment, including a guide for replication, appears in Supplemental Section \ref{sec:gpsupplement}.

Figure \ref{fig:dst-sols} gives the number of replicates that produced a successful SignalGP program for each tag-matching metric on the directional-signal task.

\subsubsection{Hamming and Streak Metrics}

Again, the hamming and streak metrics performed significantly better than all other metrics (Fisher's exact with a Holm correction for multiple comparisons, $p < 0.0008$).
We observed no significant difference in solution count between the hamming and streak metrics, however.

As in the changing-signal task, we assessed whether the streak metric produced solutions in fewer generations than the hamming metric, running 200 new replicates of each condition until 100 replicates produced a solution and recorded the number of generations that elapsed (Figure \ref{fig:dst-times}).
Among this subset of replicates, we found significantly faster generations-to-solution under the streak metric compared to the hamming metric (Wilcoxon rank-sum test, $p < 0.0016$).

\subsubsection{Integer and Hash Metrics}

As in the changing-signal task, we observed no difference in success between the integer and bidirectional integer metrics on both the changing- and directional-signal tasks.
Again, the hash metric outperformed both the integer metrics ($p < 3\times10^{-5}$).

\section{Discussion}

We used geometrical analyses to explore how tag-matching metrics constrain patterns of connectivity between tags, making some configurations unlikely or even impossible.
The bidirectional integer metric exhibited the tightest geometrical constraint in our analyses.
The unidirectional integer metric also exhibited tight geometrical constraint, but quirks of its non-commutative construction can allow that constraint to split across perfect- and worst-matching extremes.
Hamming and streak metrics exhibited looser geometric constraint, with the streak metric allowing for edge cases that very strongly break constraints.
Finally, the hash metric exhibited no geometrical constraint.

Next, we analyzed the effect of bitwise mutation on match distance score under the different metrics.
Under the hamming metric, all mutations have small effects on match distance score.
In contrast, under the integer metrics, rare mutations can have strong effects on match distance score.
The streak metric also exhibited strong-effect mutations, particularly with respect to coupling loosely-affiliated tags.
The hash metric exhibited the fattest tails of mutational magnitude, with strong-effect mutations occurring frequently.
Interestingly, the hash metric also exhibited sign-outcome frequencies that differed from the other metrics: mutations that decoupled tightly-matching tags and mutations that coupled loosely-matching tags were more frequent compared to other metrics.

The hamming metric exhibited the greatest robustness to mutation along mutational walks, followed by the streak metric.
The integer metrics, in particular the unidirectional integer metric, exhibited less robustness.
The streak metric, where all one-step mutations scramble match distance, exhibited the least robustness.

In evolutionary experiments, we found that network constraint (the number of tags a query or operand needs to simultaneously establish affinity with) influenced the relative performance of tag-matching metrics.

In target-matching evolutionary experiments, we found that the hash metric enabled rapid adaptive evolution toward targets with low network constraint.
This rapid evolution may be due to the hash metric's ability to rapidly generate variation.
Under high network constraint, however, the hash metric yielded poor-quality solutions.
The integer metrics also yielded poor-quality solutions for target graphs with network constraint.
In some more-constrained cases, the streak metric enabled more rapid adaptive evolution than the hamming metric.

In genetic programming evolutionary experiments, we found that the hamming and streak metrics yielded successful solutions the most frequently.
On the directional signal task, which tends to require denser interaction networks, we found evidence that the streak metric enabled more rapid adaptive evolution than the hamming metric.

The hash metric had the next best performance in SignalGP experiments, yielding more solutions than the integer metrics, which performed comparably.
Although the hash metric performed best in low-constraint target-matching experiments, it was outperformed in low-constraint SignalGP experiments. 
This may be due to the presence of duplication and differentiation processes across SignalGP lineages, where instruction and module count can grow over time. 

\begin{figure}
\begin{center}

\includegraphics[width=0.5\columnwidth]{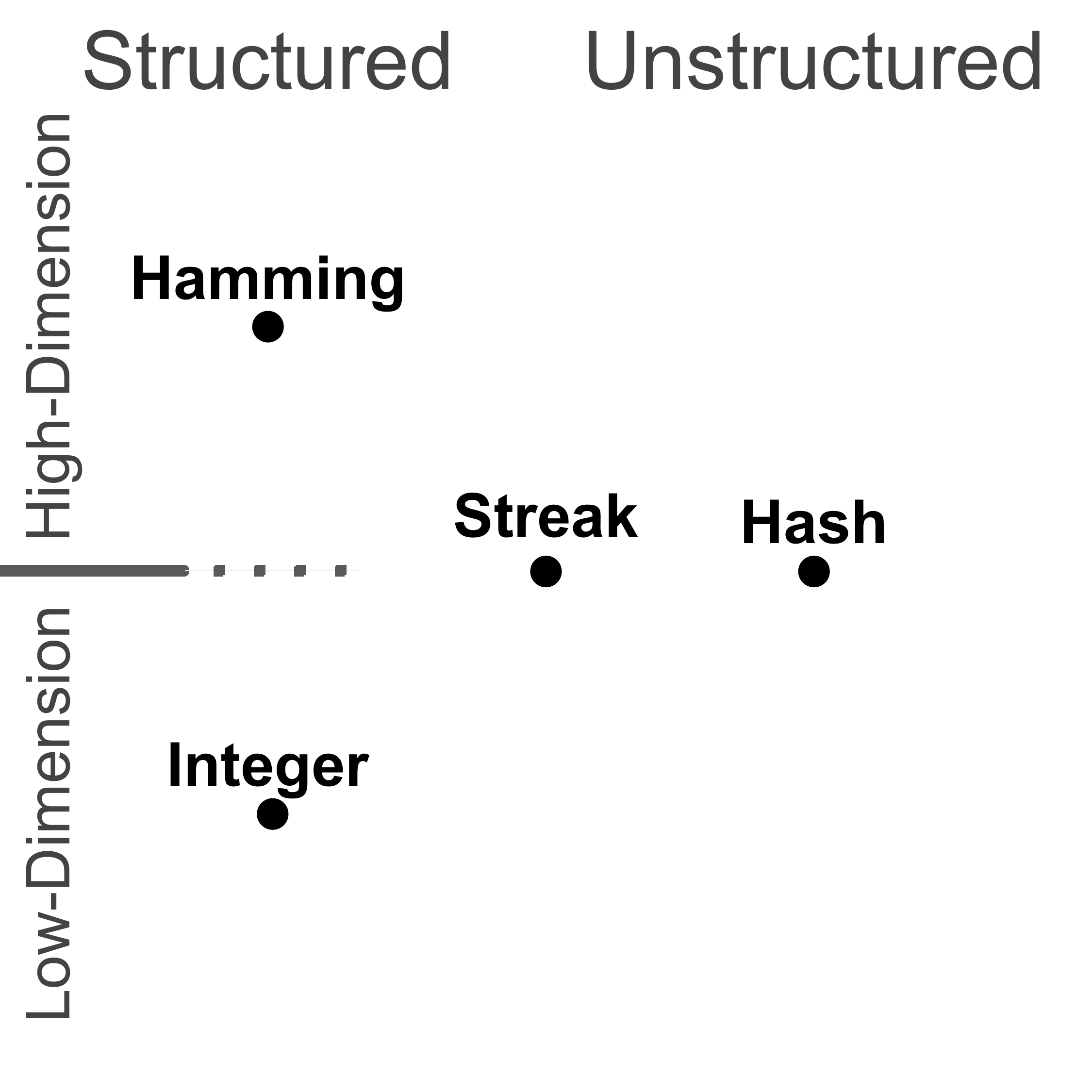}
\caption{
A conceptual schematic of the tag-matching metrics' geometric properties.
}
\label{fig:conceptual_geometry}

\end{center}
\end{figure}
 
Relative to the other metrics, the streak metric tends to offer intermediate variational and geometric properties.
Figure \ref{fig:conceptual_geometry} depicts a schematic summary of this observation.
It exhibits some, but not strict, geometric constraint.
Many mutations are neutral or near-neutral (like the integer and hamming metrics) but a fat tail of extreme-effect mutations also occur (like the hash metric).
The streak metric exhibits robustness under mutational walks that falls between the hamming and integer metrics.
These mechanistic observations offer a potential explanation for the streak metric's strong performance facilitating adaptive evolution under high-constraint conditions.
However, whether these mechanistic explanations are sufficiently complete --- especially with respect to the streak metric's outperformance of the hamming metric under high-constraint conditions --- is unclear.

\section{Conclusion}

Better understanding the mechanistic properties and functional implications of tag-matching criteria will help researchers more effectively incorporate tag matching in evolutionary systems and better understand the biases imposed by those criteria.
Within genetic programming, bespoke tag-matching criteria might increase the rate of adaptive evolution and evolving better-quality solutions.
Likewise, within artificial life bespoke tag-matching criteria might improve generation of novelty and complexity.
There has been interest, in particular, in the potential for tag-based referencing to facilitate inter-species interactions in digital ecologies \citep{dolson2021review}.

Our analyses suggests that network constraint is key to the interaction between a tag-matching scheme and problem domain.
Applications where queries much match tightly with multiple operands require high-dimensional tag-matching criteria.

The surprisingly strong performance of the hash metric on low constraint toy problems   underscores the role of tag-matching criteria in facilitating generation of phenotypic variation.

Important open questions remain with respect tag-matching criteria.
In particular, the relationships between tag-matching criteria and specificity, modularity, robustness, and the process of duplication and divergence should be explored.
Evolvability or information-theoretical analyses may prove fruitful in this regard \citep{tarapore2015evolvability}.
How to systematically design new tag-matching metrics with desirable evolutionary properties also remains an open problem.
We also need algorithms capable of computationally-efficient look ups against large sets of referents under high-dimensional or irregular tag-matching metrics. 

Tag-like mechanisms play a central role mediating interaction and function across the spectrum of biological scale \citep{holland2012signals}.
By shining light on previously-unexplored mechanistic and evolutionary properties of tagging systems, we hope that insight into artificial tag models will translate into a more nuanced appreciation of natural systems.

\section*{Funding}

This research was supported in part by NSF grants DEB-1655715 and DBI-0939454.
This material is based upon work supported by the National Science Foundation Graduate Research Fellowship under Grant No. DGE-1424871.

\section*{Conflict of interest}

The authors declare that they have no conflict of interest.

\section*{Availability of data and material}

The code used to perform and analyze our experiments, our figures, and data from our experiments is available via the Open Science Framework at \url{https://osf.io/gw5mc/} \citep{foster2017open}.
Data and figures are also organized alongside related manuscript source files at \url{https://github.com/mmore500/tag-olympics-writeup}. 

\section*{Code availability} 

We implemented our experimental systems using the Empirical library for scientific software development in C++, available at \url{https://github.com/devosoft/Empirical} \citep{charles_ofria_2019_2575607}.
Software written for these experiments is available at \url{https://github.com/amlalejini/Exploring-tag-matching-metrics-in-SignalGP/tree/1.0} and \url{https://github.com/mmore500/tag-olympics}.

\section*{Authors' contributions} Matthew Andres Moreno and Alexander Lalejini contributed to the study conception and design.
Material preparation, data collection and analysis of genetic programming benchmarks were performed by Alexander Lalejini.
Other material preparation, data collection and analysis of genetic programming benchmarks were performed by Matthew Andres Moreno.
The first draft of the manuscript was written by Matthew Andres Moreno and Alexander Lalejini and all authors commented on previous versions of the manuscript.
All authors read and approved the final manuscript.
 
\begin{acknowledgements}

Thanks to members of the DEVOLAB, in particular Nathan Rizik for help developing our tag-matching software infrastructure.
This research was supported by Michigan State University through the computational resources provided by the Institute for Cyber-Enabled Research.
This material is based upon work supported by the National Science Foundation Graduate Research Fellowship under Grant No. DGE-1424871.
Any opinions, findings, and conclusions or recommendations expressed in this material are those of the author(s) and do not necessarily reflect the views of the National Science Foundation.

\end{acknowledgements} 
\footnotesize
\bibliographystyle{apalike}
\bibliography{bibl} 

\clearpage
\newpage

\appendix

\setcounter{secnumdepth}{2}

\begin{figure}
\centering
\includegraphics[width=\linewidth]{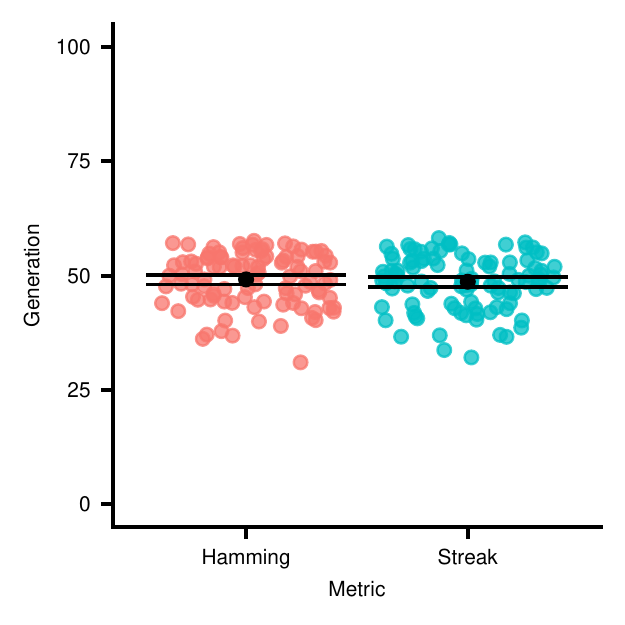}\caption{
Generations to solution for the first 100 replicates out of 200 to produce a complete solution to the changing-signal task.
Error bars indicate bootstrapped 95\% confidence intervals.
 }
\label{fig:cst-times}
\end{figure}
 
\begin{figure*}
\begin{minipage}{6in}
\includegraphics[width=\textwidth]{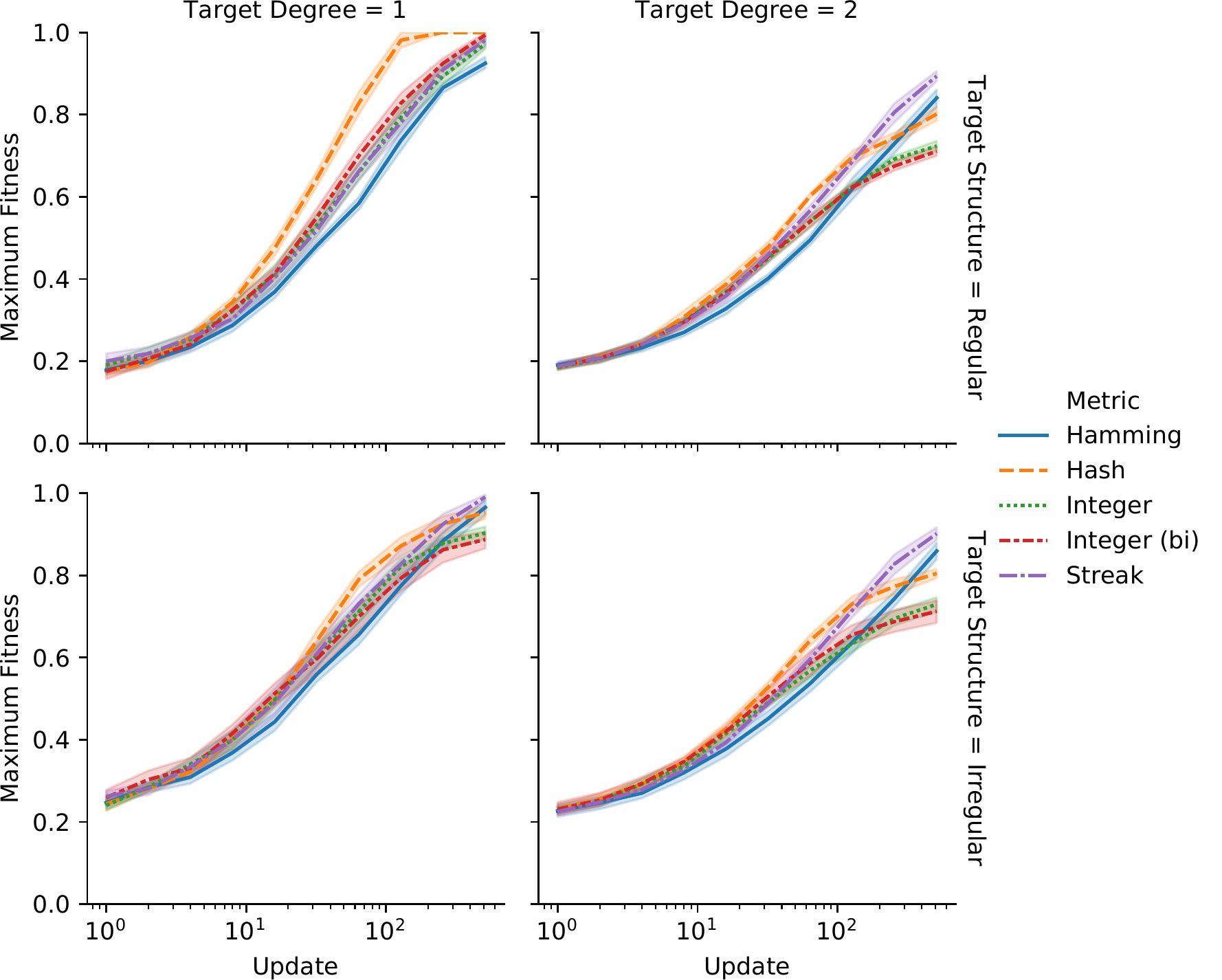}
\caption{64-node target graph}
\label{fig:evolve_big_bests}
\end{minipage}
\begin{minipage}{\textwidth}
\caption{
Trajectories of adaptive evolution for each tag-matching metric on the 64-node graph-matching task.
Maximum fitness represents the best fitness value for any individual within a population.
Here report using each metric's best-performing per-bit mutation rate.
(See Supplementary Figure \ref{fig:evolve_big_mutsweep} for survey showing how mutation rate affects adaptive evolution under each metric.)
Note log-scale x-axes.
Shaded area represents bootstrapped 95\% confidence intervals across 20 replicate observations.
}
\label{fig:evolve_bests64}
\end{minipage}
\end{figure*}

\begin{sidewaysfigure}
\vspace{100ex}
\begin{center}
\includegraphics[width=\textwidth]{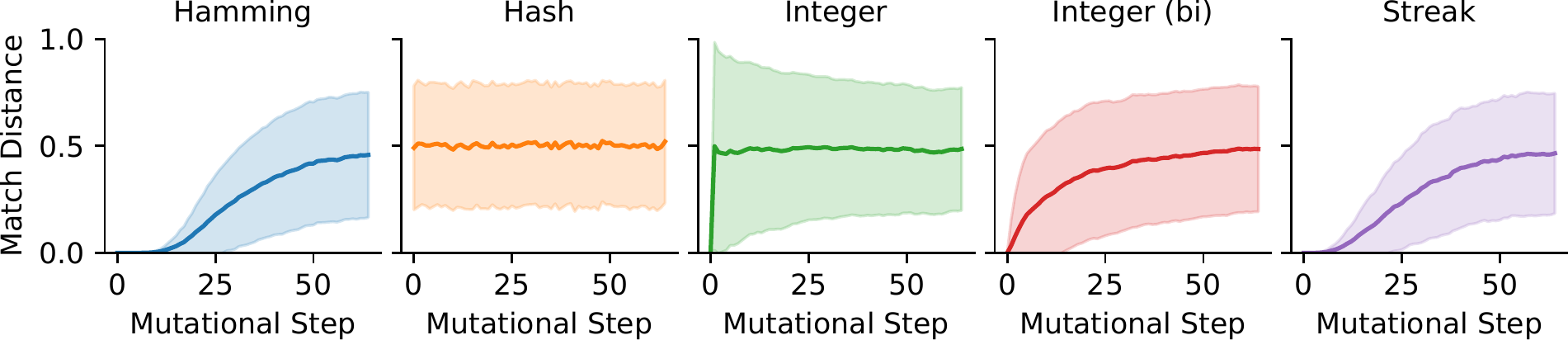}
\caption{
Match distance over mutational walks from identical tags.
Shaded area represents standard deviation.
}
\label{fig:mutational_walk_lineplot}

\end{center}
\end{sidewaysfigure}
 
\begin{figure}
\begin{center}

\begin{subfigure}{\textwidth}
\includegraphics[width=\textwidth]{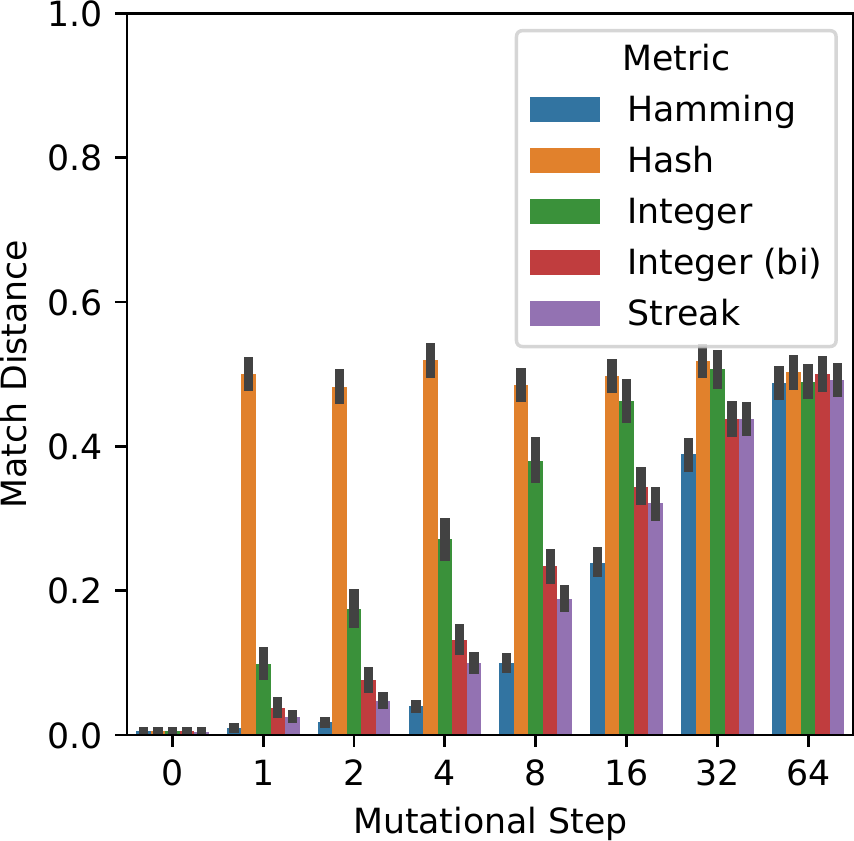}
\caption{
Error bars represent 95\% confidence intervals.
Note logarithmic scale on the $x$ axis.
}
\label{fig:mutational_walk_sampled_start_barplot}
\end{subfigure}

\begin{subfigure}{\textwidth}
\includegraphics[width=\textwidth]{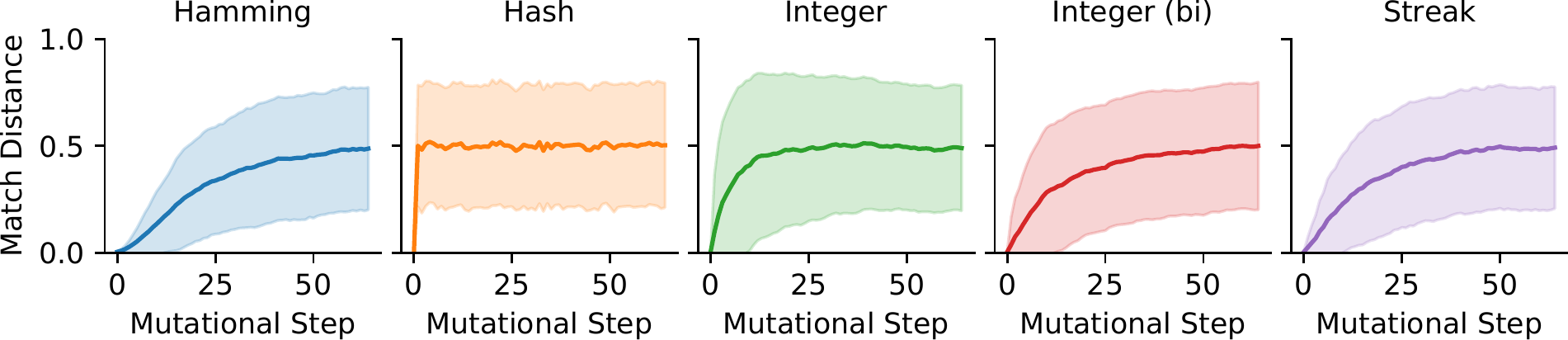}
\caption{
Alternate visualization, shaded area represents standard deviation.
}
\end{subfigure}

\caption{
Match distance along mutational walks from 32-bit tags sampled for initial match distance $<0.01$.
}
\label{fig:mutational_walk_sampled_start}

\end{center}
\end{figure}
 
\begin{figure}
\begin{center}

\begin{subfigure}{\textwidth}
\includegraphics[width=\textwidth]{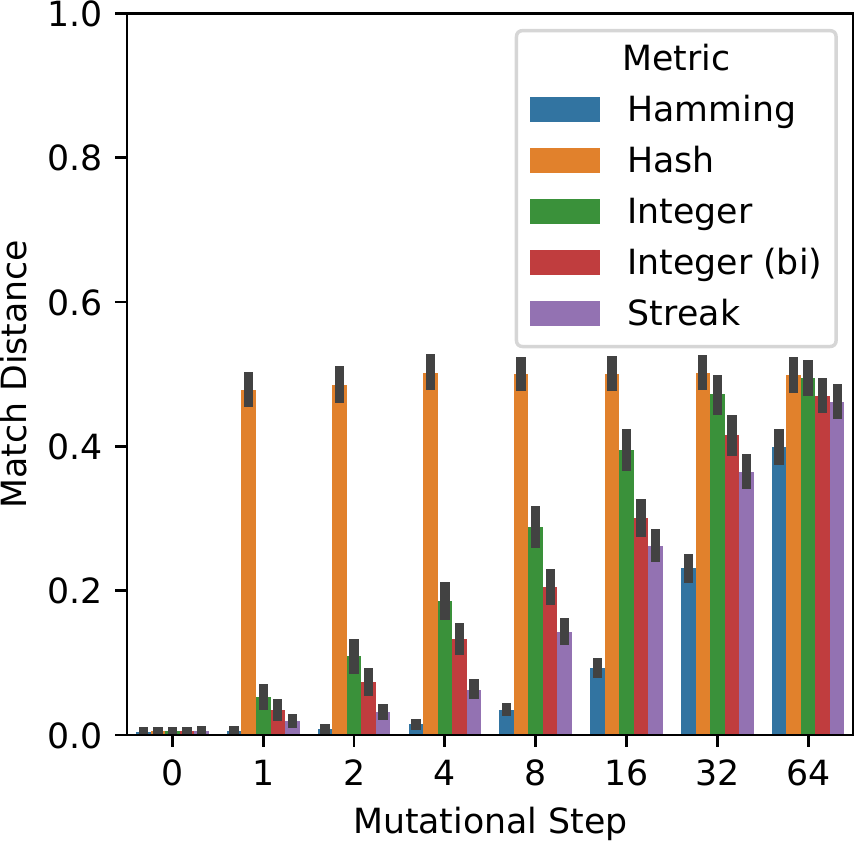}
\caption{
Error bars represent 95\% confidence intervals.
Note logarithmic scale on the $x$ axis.
}
\label{fig:mutational_walk_sampled_start_64_barplot}
\end{subfigure}

\begin{subfigure}{\textwidth}
\includegraphics[width=\textwidth]{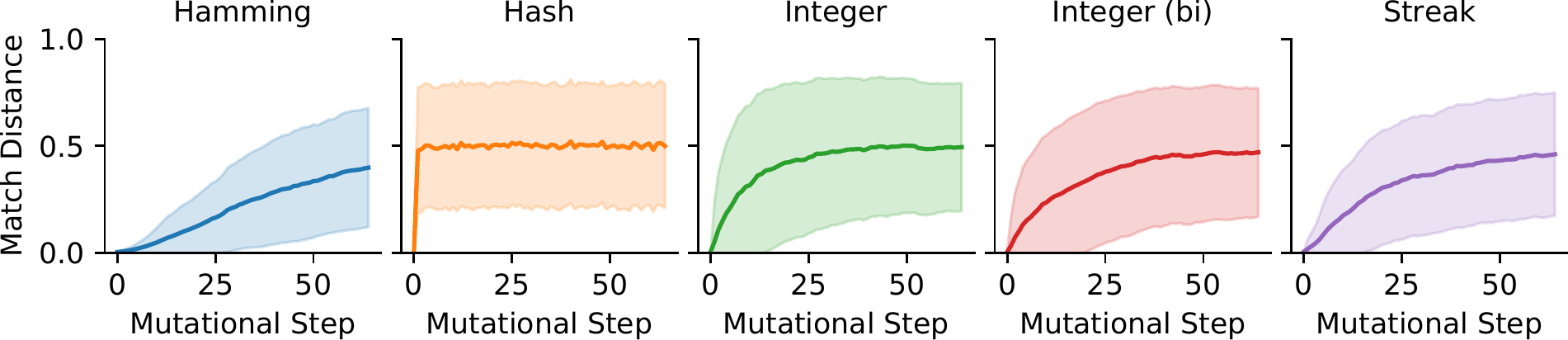}
\caption{
Alternate visualization, shaded area represents standard deviation.
}
\end{subfigure}

\caption{
Match distance along mutational walks from 64-bit tags sampled for initial match distance $<0.01$.
}
\label{fig:mutational_walk_sampled_start_64}

\end{center}
\end{figure}
 
\begin{figure*}
\begin{minipage}{6in}
\begin{center}

\includegraphics[width=\textwidth]{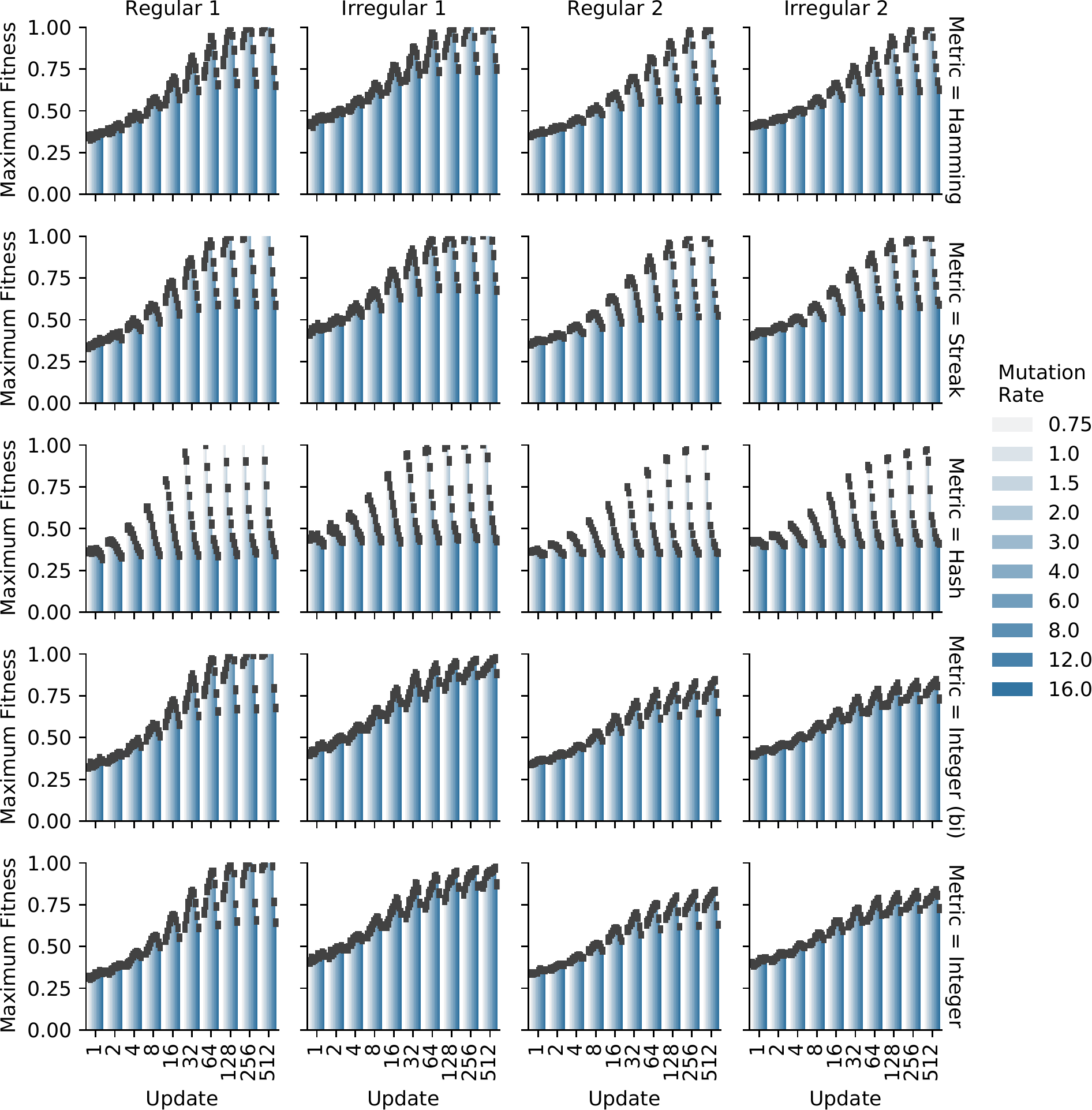}
\caption{
32-node graph-matching task mutation rate sensitivity analysis.
Metrics exhibited fastest adaptive evolution within the range of mutation rates surveyed, except the hash metric which exhibited fastest adaptive evolution at at the lowest mutation rate surveyed. 
Maximum fitness represents the best fitness value for any individual within a population.
Maximum fitness at each update is presented across the range of surveyed mutation rates.
Error bars represent bootstrap 95\% confidence intervals across 20 replicate populations.
}
\label{fig:evolve_mutsweep}

\end{center}
\end{minipage}
\end{figure*}
 
\begin{figure*}
\begin{minipage}{6in}
\begin{center}

\includegraphics[width=\textwidth]{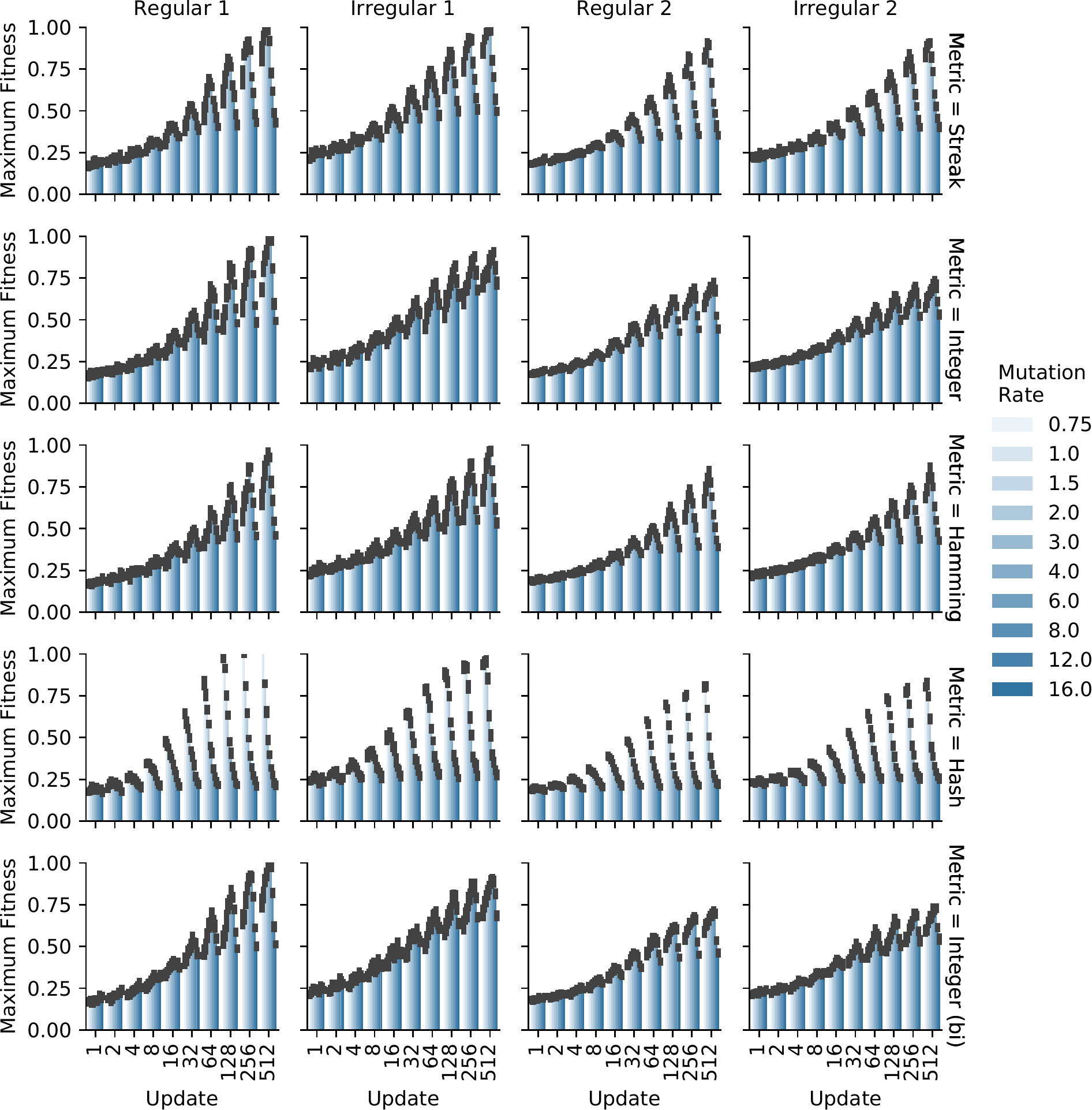}
\caption{
64-node graph-matching task mutation rate sensitivity analysis.
Metrics exhibited fastest adaptive evolution within the range of mutation rates surveyed, except the hash metric which exhibited fastest adaptive evolution at at the lowest mutation rate surveyed. 
Maximum fitness represents the best fitness value for any individual within a population.
Maximum fitness at each update is presented across the range of surveyed mutation rates.
Error bars represent bootstrap 95\% confidence intervals across 20 replicate populations.
}
\label{fig:evolve_big_mutsweep}

\end{center}
\end{minipage}
\end{figure*}
 
\section{Hamming Metric} \label{sec:hammingmetric}

Each tag was represented as an ordered, fixed-length bitstring,
\begin{align*}
t = \langle t_0, t_1, t_2, \dots, t_{n-2}, t_{n-1} \rangle
\end{align*}
where
\begin{align*}
\forall i, t_i \in \{0, 1\}.
\end{align*}

This metric is based on the work of \citep{lalejini2019else}, originally after TODO hamming cite(?).

In this metric, we compare tags according to their bitwise hamming distance.
Mathematically speaking for tags $t$ and $u$ we compute the distance according to the metric $M$ as,
\begin{align*}
M(t, u)
= \frac{
  \#\{ i : t_i \neq u_i, i=0, \dots ,n-1\}
}{
  n
}
\end{align*}

This metric is commutative and $n$-dimensional.

\section{Hash Metric} \label{sec:hashmetric}

This metric is original to the our paper and meant to serve as a control.

The an arbitrary, but determinsitic value, uniformly distributed between 0 and 1.

We rely on the \texttt{hash\_combine} function, adapted from BOOST (TODO cite).

for two values \texttt{v1} and \texttt{v2}, \texttt{hash\_combine} is defined as follows
\begin{verbatim}
unsigned int hash_combine(
  unsigned int v1,
  unsigned int v2
) {
  return v1 ^ (
    v2 * 0x9e3779b9
    + (v1 << 6) + (v1 >> 2)
  );
}
\end{verbatim}

We compute the hash value of a tag as follows
\begin{verbatim}
unsigned int h(unsigned char *tag) {
  unsigned int result = tag[0];
  for (int i = 1; i < 4; ++i){
    result = hash_combine(result, t[i]);
  }
  return result;
}
\end{verbatim}
where \texttt{tag} is the tag's bitstring stored as an array of bytes.

To compute the metric $H$ we then call \texttt{hash\_combine} to combine the hash values of the tags $t$ and $u$
\begin{align*}
H(t, u) = \texttt{hash\_combine( h(}t\texttt{), h(}u\texttt{))}
\end{align*}

Note that this is not commutative.

\section{Integer Metric}
\label{sec:integermetric}

Each tag was represented as an ordered, fixed-length bitstring,
\begin{align*}
t = \langle t_0, t_1, t_2, \dots, t_{n-2}, t_{n-1} \rangle
\end{align*}
where
\begin{align*}
\forall i, t_i \in \{0, 1\}.
\end{align*}

This metric is inspired by \citep{spector2011tag}.
They used positive integers between 0 and 100 to name referents.
Queries were provided the referent that had the next-larger value, wrapping around from 100 back to 0.

In this metric, we compare tags according to their value as an unsigned integer according to the following representation $f$,
\begin{align*}
f(t)
= \sum_{i=0}^{n-1} t_i \times 2^i.
\end{align*}

The distance metric $I$ between two length-$n$ tags $t$ and $u$ is
\begin{align*}
I(t, u) =
\begin{cases}
  \frac{2^n - f(t) + f(u)}{2^n}, & \text{if } f(t) > f(u), \\
  \frac{f(t) - f(u)}{2^n},         & \text{else} f(t) \leq f(u).
\end{cases}
\end{align*}

Note that this metric is non-commutative, i.e., it is not necessarily true that $I(t, u) = I(u, t)$.

Note also that this metric is one-dimensional.

A algorithmic advantage of this metric is that it allows for log-time matching.

\subsection{Bidirectional Integer Metric} \label{sec:integerbimetric}

Each tag was represented as an ordered, fixed-length bitstring,
\begin{align*}
t = \langle t_0, t_1, t_2, \dots, t_{n-2}, t_{n-1} \rangle
\end{align*}
where
\begin{align*}
\forall i, t_i \in \{0, 1\}.
\end{align*}

This metric is inspired by \citep{spector2011tag}.
They used positive integers between 0 and 100 to name referents.
Queries were provided the referent that had the next-larger value, wrapping around from 100 back to 0.

In this metric, we compare tags according to their value as an unsigned integer according to the following representation $f$,
\begin{align*}
f(t)
= \sum_{i=0}^{n-1} t_i \times 2^i.
\end{align*}

The distance metric $I$ between two length-$n$ tags $t$ and $u$ is
\begin{align*}
I(t, u) =
\begin{cases}
  \frac{2^n - f(t) + f(u)}{2^n}, & \text{if } f(t) > f(u), \\
  \frac{f(t) - f(u)}{2^n},         & \text{else} f(t) \leq f(u).
\end{cases}
\end{align*}

Note that this metric is non-commutative, i.e., it is not necessarily true that $I(t, u) = I(u, t)$.

Note also that this metric is one-dimensional.

\section{Streak Metric} \label{sec:streakmetric}

Each tag was represented as an ordered, fixed-length bitstring,
\begin{align*}
t = \langle t_0, t_1, t_2, \dots, t_{n-2}, t_{n-1} \rangle
\end{align*}
where
\begin{align*}
\forall i, t_i \in \{0, 1\}.
\end{align*}

This metric was originally proposed by \citep{downing2015intelligence}.
Downing claims that it exhibits
It is computed according to the ratio between the longest contiguously matching substring among two bitsets and the longest contiguously mismatching substring among those two bitsets.
Downing claims that this metric exhibits greater robustness compared to integer and hamming distance metrics using mutational walk experiments but does not demonstrate it in an evolving system.

We define the greatest contiguously-matching length of $n$-long bitstrings $t$ and $u$ as follows,
\begin{align*}
m(t, u) = \max(\{i - j \forall i, j \in 0..n-1 \mid \forall q \in i..j, t_q = u_q \})
\end{align*}

We define the greatest contiguously-mismatching length as follows,
\begin{align*}
n(t, u) = \max(\{i - j \forall i, j \in 0..n-1 \mid \forall q \in i..j, t_q \neq u_q \})
\end{align*}

The streak metric $S'$  with tags $t$ and $u$
\begin{align*}
S'(t, u)
= \frac{ p(n(t,u)) }{p(m(t,u)) + p(n(t,u))}.
\end{align*}
where $p$ approximates the probability of a contiguously-matching substring between

It is worth noting that the formula for computing the probability of a $k$-bit match or mismatch, given by Downing as follows, is actually mathematically flawed.
\begin{align*}
p_k
= \frac{n - k + 1}{2^k}
\end{align*}

The probability of a $0$-bit match according to this formula would be computed as $p_0 = \frac{n - 0 + 1}{2^0} = n + 1$ which is clearly impossible because $p_0 > 1 \forall n > 0$.
The actual can probability be achieved using a lookup table computed using dynamic programming.

However, the formula Downing presented provides a useful approximation to the probability of a $k$ bit match.
For computational efficiency and consistency with the existing literature we use clamp edge cases between 0 and 1 to yield the corrected streak metric $S$.

\begin{align*}
S(t, u) =
\max( \min( S'(t, u), 1), 0)
\end{align*}

To get a sense of the regularity, in a looses sense, of each space we uniformly sampled triplets of points $A$, $B$, and $C$.
Then, for each metric $m$ we calculated the statistic $m(A, B) + m(B, C) - m(A, C)$.
If the triangle inequality is respected this statistic should be greater than or equal to zero.
Figure \ref{fig:detour_difference} plots the distribution of this statistic for each metric.
The hamming, hash, and streak metrics show evidence of ``shortcuts'' that violate the triangle inequality.
It should be noted that the raw hamming metric does respect the triangle inequality.

\section{Genetic Programming Experiments}
\label{sec:gpsupplement}

\subsection{SignalGP}

SignalGP (Signal-driven Genetic Programs) is a GP representation that enables signal-driven (\textit{i.e.}, event-driven) program execution.
In SignalGP, programs are segmented into modules (functions) that may be automatically triggered by exogenously- or endogenously-generated signals.
Each module in SignalGP associates a tag with a linear sequence of instructions.
SignalGP makes explicit the concept of signals (events), which comprise a tag and, optionally, signal-specific data.
Signals trigger the module with the closest matching tag (according to a given tag-matching scheme), using any signal-associated data as input to the triggered module.
SignalGP can handle many signals simultaneously, processing each in parallel.

The SignalGP instruction set, in addition to including traditional GP operations, allows programs to generate internal signals, broadcast external signals, and otherwise work in a tag-based context.
Instructions contain arguments, including an evolvable tag, that may modify the instruction's effect, often specifying memory locations or fixed values.
Instructions may refer to program modules using tag-based referencing; for example, an instruction may trigger the execution of a program module using the instruction's tag to specify which module to trigger.
SignalGP also supports genetic regulation with promoter and repressor instructions that, when executed, allow programs to adjust how well subsequent signals match with a target function (specified with tag-based referencing).

See \citepinappendix{lalejini2018evolving} for a more detailed description of the SignalGP representation. Additionally, see the GitHub repository for the SignalGP implementation used in this work \citepinappendix{lalejini_2020_3781295}.

\subsection{Changing-signal Task Description}

The changing-signal task requires programs to express a distinct response to each of $K$ environmental signal (each signal has a unique tag).
Programs express a response by executing one of $K$ response instructions.
Successful programs can `hardcode' each response with the appropriate environmental signal, ensuring that each environmental signal's tag best matches the function containing the correct response.
We expect the particular metric used to match tags to influence how well programs adapt to changing-signal task.

During evaluation, we afford programs 64 time steps to express the appropriate response after receiving a signal.
Once a program expresses a response or the allotted time expires, we reset the program's virtual hardware (resetting all executing threads and thread-local memory), and the environment produces the next signal.
Evaluation continues until the program correctly responds to each of the $K$ environmental signals or until the program expresses an incorrect response.
During each evaluation, programs experience environmental signals in a random order; thus, the correct \textit{order} of responses will vary and cannot be hardcoded.

For each metric, we evolved 200 replicate populations (each with a unique random number seed) of 500 asexually reproducing programs in an eight-signal environment ($K=8$) for 100 generations.
We identified the most performant tag mutation rate (from a range of possible mutation rates) for each metric to use in our experiment.
These data (and analyses) are available online in the GitHub repository that houses these experiments \citepinappendix{lalejini_2020_3781295}.
We used the following per-bit tag mutation rates for the changing-signal task: 0.01 for the Hamming and Streak metrics, 0.002 for the Hash metric, and 0.02 for the Integer and Bidirectional Integer metrics.
Aside from tag mutation rate, the overall configuration used for each metric was identical.
We limited tag variation in offspring to tag mutation operators (bit flips) by initializing populations with a common ancestor program in which all tags are identical and by disallowing mutations that would insert instructions with random tags.

The full configuration details for the changing-signal task (including a guide to running these experiments on your local machine) can be found in the associated GitHub repository \citepinappendix{lalejini_2020_3781295}.

\subsection{Directional-signal Task Description}

As in the changing-signal task, the directional-signal task requires that programs respond to a sequence of environmental cues; in the directional-signal task, however, the correct response depends on previously experienced signals.
In the directional signal task, there are two possible environmental signals --- a `forward-signal' and a `backward-signal' (each with a distinct tag) ---  and four possible responses.
If a program receives a forward-signal, it should express the next response, and if the program receives, a backward-signal, it should express the previous response.
For example, if response-three is currently required, then a subsequent forward-signal indicates that response-four is required next, while a backward-signal would instead indicate that response-two is required next.
Because the appropriate response to both the backward- and forward-signals change over time, successful programs must regulate which functions these signals trigger (rather than hardcode each response to a particular signal).

We evaluate programs on all possible four-signal sequences of forward- and backward-signals (sixteen total).
For each program, we evaluate each sequence of signals independently, and a program's fitness is equal to its aggregate performance.
Otherwise, evaluation on a single sequence of signals mirrors that of the changing signal task.

We used an identical experimental design for the directional-signal task as in the changing-signal task.
However, we evolved programs for 5000 generations (instead of 100) and re-parameterized each metric's tag mutation rate (these data are available in the associated GitHub repository \citepinappendix{lalejini_2020_3781295}):
0.001 for the Hamming and Hash metrics, 0.002 for the Integer and Streak metrics, and 0.0001 for the Bidirectional Integer Metric.

The full configuration details for the directional-signal task (including a guide to running these experiments on your local machine) can be found in the associated GitHub repository \citepinappendix{lalejini_2020_3781295}.

\subsection{Data analysis and Implementation}

The source code for our GP experiments can be found in the following GitHub repository: \citepinappendix{lalejini_2020_3781295}. This repository additionally includes all data analysis and visualization scripts, experiment configuration details, and a guide for running our experiments locally.
  
\footnotesize
\bibliographystyleinappendix{apalike}
\bibliographyinappendix{bibl} 

\end{document}